\documentclass[10pt,twocolumn,letterpaper]{article}

\usepackage{cvpr}      
\definecolor{cvprblue}{rgb}{0.21,0.49,0.74}
\usepackage[pagebackref,breaklinks,colorlinks,allcolors=cvprblue]{hyperref}
\usepackage{multirow}
 
\usepackage[table]{xcolor}
\newcommand{\std}[1]{{\scriptsize $\pm$ #1}}
\newcommand{\mysection}[1]{\vspace{1.0pt}\noindent\textbf{#1}}

\def\paperID{8717} 
\def\confName{CVPR}
\def\confYear{2026}

\title{AdaSpot: Spend Resolution Where It Matters for Precise Event Spotting}

\author{Artur Xarles$^{1,2}$\hspace{0.7cm} Sergio Escalera$^{1,2,3}$\hspace{0.7cm} Thomas B. Moeslund$^{3}$\hspace{0.7cm} Albert Clapés$^{1,2}$ \\
$^{1}$Universitat de Barcelona, Barcelona, Spain\\
$^{2}$Computer Vision Center, Cerdanyola del Vallès, Spain \\
$^{3}$Aalborg University, Aalborg, Denmark\\
{\tt\small arturxe@gmail.com}, {\tt\small sescalera@ub.edu}, {\tt\small tbm@create.aau.dk}, {\tt\small aclapes@ub.edu} \\
}

\begin{document}
\maketitle

\begin{abstract}
Precise Event Spotting aims to localize fast-paced actions or events in videos with high temporal precision, a key task for applications in sports analytics, robotics, and autonomous systems. Existing methods typically process all frames uniformly, overlooking the inherent spatio-temporal redundancy in video data. This leads to redundant computation on non-informative regions while limiting overall efficiency. To remain tractable, they often spatially downsample inputs, losing fine-grained details crucial for precise localization. To address these limitations, we propose \textbf{AdaSpot}, a simple yet effective framework that processes low-resolution videos to extract global task-relevant features while adaptively selecting the most informative region-of-interest in each frame for high-resolution processing. The selection is performed via an unsupervised, task-aware strategy that maintains spatio-temporal consistency across frames and avoids the training instability of learnable alternatives. This design preserves essential fine-grained visual cues with a marginal computational overhead compared to low-resolution-only baselines, while remaining far more efficient than uniform high-resolution processing. Experiments on standard PES benchmarks demonstrate that \textbf{AdaSpot} achieves state-of-the-art performance under strict evaluation metrics (\eg, $+3.96$ and $+2.26$ mAP$@0$ frames on Tennis and FineDiving), while also maintaining strong results under looser metrics. Code is available at: \href{https://github.com/arturxe2/AdaSpot}{https://github.com/arturxe2/AdaSpot}.
\end{abstract}

\section{Introduction}
\label{sec:intro}

\begin{figure}[t]
    \centering
    \includegraphics[width=\columnwidth]{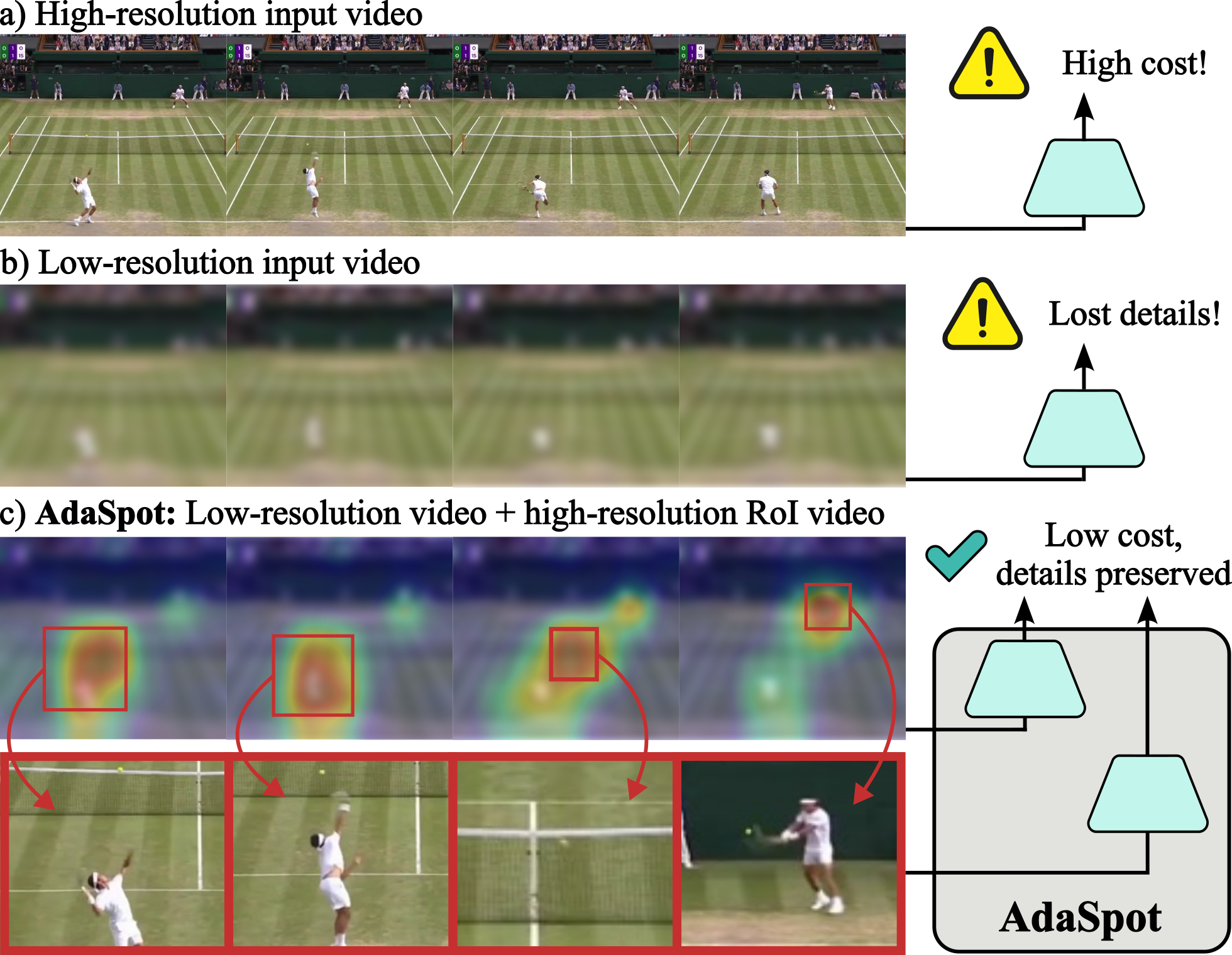}
    \caption{Illustration of standard PES approaches: (a) high-resolution videos incur high computational cost, whereas (b) low-resolution videos reduce cost but lose fine-grained details crucial for precise temporal localization. In contrast, (c) \textbf{AdaSpot} captures global context from low-resolution videos and adaptively applies high-resolution processing to task-relevant regions, preserving fine-grained details efficiently.}
    \label{fig:topright}
    \vspace{-0.5cm}
\end{figure}

Recent progress in action recognition~\cite{kong2022human} has enabled reliable classification of \textit{what} happens in a video. However, many applications also require awareness of \textit{when} it happens. Precise temporal detection --\ie, determining \textit{exactly when} an action or event occurs-- is crucial for tasks such as identifying decisive sports moments~\cite{rafiq2020scene, naik2022comprehensive}, anticipating pedestrian behavior~\cite{sighencea2021review}, and facilitating responsive human–robot interaction~\cite{zhdanova2020human, keshinro2023human}. Within this context, two established formulations are Temporal Action Localization (TAL)~\cite{xia2020survey} and Event Spotting (ES)~\cite{xu2025action}. TAL models actions as temporal segments, whereas ES represents an action or event using a single keyframe. This representation makes ES well-suited for fast scenarios where events are brief. Precise Event Spotting (PES)~\cite{hong2022spotting, mcnally2019golfdb} further refines ES by enforcing near frame-level accuracy, therefore increasing the task's difficulty, as even minor temporal errors can result in missed events. Although sports datasets currently dominate PES benchmarks due to their fast-paced nature and the need for high temporal precision, PES itself is domain-agnostic and broadly applicable to any setting where accurate temporal detection is critical.

Existing PES methods primarily focus on temporal modeling, exploring multi-scale representations and long-range dependencies~\cite{xarles2024t, xarlest, santra2025precise}. However, they typically process all frames uniformly, disregarding the substantial spatio-temporal redundancy inherent in videos. This uniform processing leads to high computational costs on high-resolution inputs (\cref{fig:topright}(a)), as much computation is spent on regions with limited task relevance. To remain tractable, models are often trained on spatially downsampled videos (\cref{fig:topright}(b)), while keeping high temporal resolution to meet the task's precision requirements. Yet, spatial downsampling can cause the loss of fine details observable only at high resolutions --details that are crucial for precise temporal detection (\eg, in tennis, the subtle cue of the ball contacting the ground can vanish, hindering exact frame identification). This issue is further amplified in far-view scenes, where action cues occupy only a small portion of the frame.

Prior work in video action recognition has addressed similar challenges through dynamic computation strategies~\cite{han2021dynamic} that adaptively allocate computational resources to task-relevant regions. A prominent line of work~\cite{adafocus-v1, adafocus-v2, adafocus-v3, adafocus-v4, zheng2023dynamic, liu2022task} focuses on reducing spatial redundancy at the input level by first employing lightweight modules to identify informative regions, which are then processed with higher-capacity computation. This approach effectively reduces unnecessary computation on uninformative areas while maintaining strong performance by concentrating resources where they matter most. However, most existing methods rely on learnable cropping mechanisms~\cite{adafocus-v1, adafocus-v2, adafocus-v3, zheng2023dynamic} for region selection, which can be unstable to train even in standard action recognition settings~\cite{adafocus-v4}. In the case of PES, where supervision signals are weaker due to the highly localized spatio-temporal nature of events, directly applying such cropping-based approaches amplifies these instabilities, often leading to inconsistent or unreliable crops across frames (see \cref{sec:redundancy}).

To address these limitations while effectively mitigating spatial redundancy in PES, we propose \textbf{AdaSpot} (\cref{fig:topright}(c)), a simple yet effective framework that adaptively focuses computation on task-relevant regions. AdaSpot operates at multiple resolutions: it first processes full frames at low resolution to extract global task-relevant features and guide the selection of a single region of interest (RoI) for each frame. These RoIs are then processed at high resolution --\ie, with increased computational capacity-- to capture fine-grained details, which are fused with the global features to preserve both local and global information. RoI selection is performed using saliency maps derived from the low-resolution features in a training-free manner, avoiding the instability of alternative learnable cropping mechanisms. Our RoI selector further addresses three key challenges when extracting RoIs from saliency maps: (1) it replaces zero-padding with replicate padding to mitigate \textit{center bias}, (2) applies spatio-temporal smoothing to reduce \textit{noisy activations} and ensure consistent RoI selection, and (3) adapts RoI size according to the saliency spread to handle \textit{varying required RoI sizes} across datasets, action types, or camera views. This design enables AdaSpot to capture fine-grained, task-relevant details at low computational cost, since only a small portion of each frame is processed at high resolution. Our main contributions can be summarized as follows:

\begin{itemize}
    \item To the best of our knowledge, we introduce the first PES framework that explicitly addresses spatial redundancy at the input level by adaptively allocating high-resolution processing only to the most task-relevant region of each frame. This design preserves fine-grained visual cues essential for frame-level precision, introducing only marginal overhead compared to a low-resolution-only baseline, while still incurring far less computational cost than uniform high-resolution processing.

    \item We propose an unsupervised, task-aware RoI selection strategy based on saliency maps, avoiding the training instability of learnable cropping alternatives. Our RoI selector mitigates activation bias and noise to ensure robust and consistent localization across frames, while dynamically adapting region sizes to varying scene and action characteristics.

    \item \textbf{AdaSpot} achieves state-of-the-art results across multiple PES benchmarks under tight temporal error tolerances, while improving --or, at least-- maintaining competitive computational efficiency relative to prior work.

\end{itemize}


\section{Related Work}
\label{sec:related_work}

ES addresses the problem of identifying when actions or events occur within a video. In current ES literature~\cite{hong2022spotting, giancola2024deep}, both extended actions and brief events are represented using a single keyframe. Following this convention, we use the terms \textit{action} and \textit{event} interchangeably, as the distinction does not affect our work.

\mysection{Event spotting.} Given their conceptual similarity, ES and TAL methods often share architectural components and can be broadly categorized into \textit{end-to-end} and \textit{two-stage} formulations. \textit{End-to-end} models~\cite{hong2022spotting, xarles2024t, xarlest, santra2025precise, tran2024unifying, denize2024comedian, liu2025few, voeikov2020ttnet} jointly learn visual feature extraction and temporal modeling within a unified framework, typically employing lightweight 2D backbones with local temporal modules to maintain training efficiency. In contrast, \textit{two-stage} approaches~\cite{shi2023tridet, zhang2022actionformer, soares2022temporally, zhou2021feature, xarles2023astra} decouple these processes, first extracting video features using larger 2D or 3D encoders, followed by a separate temporal modeling stage. Temporal modeling is commonly realized through sequential architectures such as RNNs~\cite{hong2022spotting, santra2025precise, tran2024unifying}, Transformers~\cite{soares2022temporally, zhang2022actionformer, zhou2021feature, xarles2023astra, denize2024comedian}, or Temporal Convolutions~\cite{soares2022temporally, xarles2024t, xarlest, shi2023tridet}. To capture short- and long-range dependencies, several works adopt multi-scale processing, employing either pyramid networks~\cite{zhang2022actionformer, shi2023tridet} or U-Net-like architectures~\cite{xarles2024t, xarlest, soares2022temporally}.

Recent ES research increasingly favors end-to-end pipelines following E2E-Spot~\cite{hong2022spotting}, which demonstrates that simple joint modeling can outperform multi-stage alternatives while enabling low-latency inference. Subsequent work further advances temporal modeling within this framework: T-DEED~\cite{xarles2024t, xarlest} replaces Gate Shift Modules (GSM)~\cite{sudhakaran2020gate} with Gate Shift Fuse (GSF)~\cite{sudhakaran2023gate} and introduces multi-scale temporal processing; ~\citet{santra2025precise} incorporates long-range refinement modules within the backbone; and UGLF~\cite{tran2024unifying} adds a vision-language branch to highlight semantically salient content. Despite these advances, most methods still ignore spatio-temporal redundancy and operate on downsampled inputs, losing fine-grained detail. Instead, explicitly addressing redundancy can enable adaptive computation, focusing high-resolution processing and spending additional compute only where it matters.

\mysection{Reducing spatio-temporal redundancy.} Prior work in video action classification attempts to mitigate the spatio-temporal redundancy inherent in videos by concentrating computation on task-relevant regions. These approaches can be broadly categorized as \textit{architecture-based} or \textit{input-based}. \textit{Architecture-based} methods retain full-frame processing but allocate computation unevenly across feature map locations. Examples include deformable and sparse networks ~\cite{dai2017deformable, liu2015sparse, xia2022vision, child2019generating}, which dynamically attend to salient spatial or temporal positions. However, such approaches typically still require dense early-stage processing to establish global context before selectively attending to informative regions, which constrains their efficiency gains.

\textit{Input-based} methods reduce redundancy at the input level by first identifying relevant regions using a lightweight mechanism, which are then processed at higher resolution and/or with larger networks. Early work focused on \textbf{temporal redundancy}~\cite{wu2019multi, meng2020ar, lin2022ocsampler, ghodrati2021frameexit, wu2020dynamic, korbar2019scsampler, xia2022nsnet}, either by adjusting frame resolution based on frame importance~\cite{meng2020ar}, selecting the most informative frames~\cite{lin2022ocsampler}, or stopping inference once sufficient evidence is obtained~\cite{ghodrati2021frameexit}. In PES, however, temporal redundancy is harder to deal with --skipping frames risks missing entire events. In contrast, \textbf{spatial redundancy}~\cite{karpathy2014large, jana2021unsupervised, kapidis2019egocentric, adafocus-v1, adafocus-v2, adafocus-v3, adafocus-v4, zheng2023dynamic, liu2022task} is more suitable for the nature of the task. Early strategies used naïve center cropping~\cite{karpathy2014large}, motion-based region selection~\cite{jana2021unsupervised}, or object-based region selection~\cite{kapidis2019egocentric}. More recent approaches perform task-aware region selection, as exemplified by the AdaFocus family~\cite{adafocus-v1, adafocus-v2, adafocus-v3, adafocus-v4} and CoViFocus~\cite{zheng2023dynamic}, which learn per-frame crop regions through reinforcement learning or differentiable cropping. Yet, learning crop locations directly in pixel space~\cite{adafocus-v2, zheng2023dynamic} remains challenging due to limited supervision, low input diversity, and training instability~\cite{adafocus-v4}. Learning crop positions in feature space~\cite{adafocus-v3, adafocus-v4} alleviates some of these issues; however, applying such methods to PES remains difficult (see ~\cref{sec:redundancy}) because of the weaker supervision signals associated with short and highly localized PES events. Alternatively, softer approaches such as saliency-guided warping~\cite{liu2022task} expand discriminative regions while retaining global frame structure, though they can introduce geometric distortions that hinder spatio-temporal modeling.

In the context of PES, redundancy-aware methods remain underexplored. Concretely, only UGLF~\cite{tran2024unifying} attempts to mitigate spatial redundancy architecturally by using a vision-language model to focus on features corresponding to task-relevant concepts (\eg, player or ball in football). However, it requires hand-crafted dataset-specific vocabularies, does not exploit higher-resolution for relevant regions, and offers limited efficiency gains due to the overhead of the vision-language model.

\vspace{0.15cm}
\noindent We address spatial redundancy in PES with an input-based approach. To overcome the challenges of learning reliable RoIs in prior input-based methods, we introduce an unsupervised, task-aware strategy based on saliency, which mitigates training instabilities while generating semantically meaningful and temporally consistent RoIs across frames. Within PES, our method differs from UGLF in that it operates directly on the input space for greater computational efficiency, selects task-aware regions without requiring any dataset-specific vocabularies, and avoids dependence on large vision-language models.


\section{Method}

\begin{figure*}[t]
    \centering
    \includegraphics[width=\linewidth]{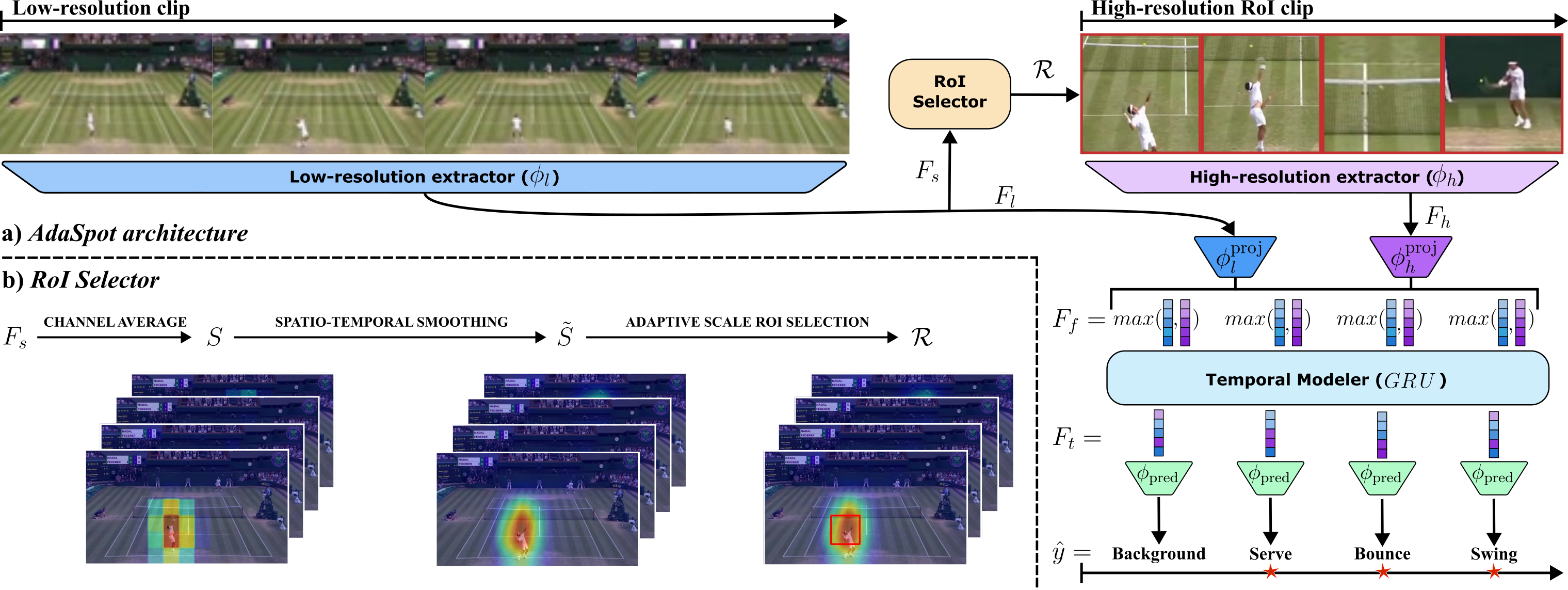}
    \caption{Overview of our proposed method, \textbf{AdaSpot}. (a) The framework uses a \textit{low-resolution extractor} to process low-resolution clips and generate global features $F_l$ and spatial maps $F_s$. A \textit{RoI selector} leverages $F_s$ to identify the most relevant region in each frame. The resulting RoI sequence is then processed by a \textit{high-resolution extractor} to capture fine-grained features, $F_h$. $F_l$ and $F_h$ are linearly projected, aggregated, and passed through a \textit{temporal modeler}, before a \textit{prediction head} produces per-frame classifications. (b) Details of the RoI selector: channel averaging generates saliency maps from $F_s$, spatio-temporal smoothing reduces noise, and adaptive-scale RoI selection adjusts the RoI size to the saliency spread.}
    \label{fig:modelFigure}
    \vspace{-0.5cm}
\end{figure*}

\mysection{Problem definition.} PES~\cite{hong2022spotting} aims to localize discrete actions or events within an untrimmed video $X$. The goal is to identify all event instances $E = \{e_1, \dots, e_N\}$, where $N$ is the number of events and may vary across videos. Each event $e_i$ is defined by an event class $c_i \in \{1, \dots, C\}$, with $C$ the total number of event categories, and a temporal position $t_i$ indicating the frame corresponding to the exact or most representative moment of that action or event. Thus, each event can be represented as a pair $e_i = (c_i, t_i)$.

\subsection{Methodology}
Our proposed method, \textbf{AdaSpot}, is designed for PES, where the model adaptively identifies the most task-relevant region in each frame and processes it at high resolution, enabling the capture of fine-grained visual cues crucial for precise temporal localization. As shown in \cref{fig:modelFigure}(a), \textbf{AdaSpot} consists of a low-resolution feature extractor, a RoI selector, a high-resolution feature extractor, a temporal modeler, and a prediction head. Videos are divided into fixed-length clips of $L$ densely sampled frames. Each clip is provided in high resolution, from which a low-resolution version is first derived and passed through the low-resolution feature extractor to: (1) capture per-frame global context features relevant to the task, and (2) produce spatially structured feature maps that guide the identification of the most informative region within each frame. The RoI selector then uses these feature maps to generate saliency maps and identify one RoI per frame, while enforcing spatio-temporal consistency across frames. The selected regions are aggregated on the fly to form a high-resolution clip of RoIs, which is processed by the high-resolution feature extractor to obtain fine-grained per-frame representations. The temporal modeler fuses features from both the low- and high-resolution branches, combining global context and local details, and refines them through a long-term temporal module to capture temporal dependencies. Finally, the prediction head classifies each frame as either an event or background.

\subsubsection{Low-resolution feature extractor}

The low-resolution feature extractor $\phi_l$ operates on the full-view input sequence $X_l \in \mathbb{R}^{L\times W_l \times H_l \times 3}$, obtained by resizing each frame to $W_l \times H_l$. Following~\citet{hong2022spotting}, we adopt RegNetY~\cite{radosavovic2020designing}, a highly efficient 2D ConvNet, as our feature extractor. To capture local temporal information, GSF~\cite{sudhakaran2023gate} modules are embedded into each of its bottleneck blocks. The extractor outputs global context features $F_l=\phi_l(X_l)\in \mathbb{R}^{L\times d}$, where $d$ is the channel dimension. We also retain the feature maps from the final layer before spatial aggregation, $F_s\in \mathbb{R}^{L\times W_s \times H_s \times d}$, which preserve spatial structure, and use them to generate the saliency maps that guide the RoI selector.

\subsubsection{RoI selector}

We propose a training-free RoI selection mechanism (\cref{fig:modelFigure}(b)) that bypasses the instability of learning-based methods while producing stable, semantically meaningful regions across frames. Specifically, it leverages the intrinsic activation patterns in the low-resolution feature maps $F_s$ to guide the selection of the most informative regions.

\mysection{Saliency map generation.} As noted by~\citet{zhou2016learning}, activation maps from deeper convolutional layers tend to exhibit stronger responses over task-relevant regions. We exploit this by averaging $F_s$ along the channel dimension to obtain saliency maps $S\in \mathbb{R}^{L\times W_s \times H_s}$. Each frame map $S_l$, $l \in \{0, \dots, L-1\}$, is then min-max normalized for consistent scaling across frames. Since $F_s$ comes from deep backbone layers, its spatial resolution is heavily downsampled. Selecting RoIs directly on this coarse grid limits RoIs to a few discrete spatial positions --so even a one-cell shift can correspond to a large displacement in the original frame, causing abrupt and unstable RoI changes. To address this, we upsample $S_l$ by a factor $k$ along the spatial dimension. While this does not add information, it provides a denser sampling grid for RoI selection, resulting in more precise localization and smoother temporal trajectories.

\mysection{Stabilizing the saliency maps.} Extracting RoIs from $S$ involves three main challenges: (1) \textit{Center bias:} zero-padding in convolutional layers can reduce activation strength near image borders~\cite{alsallakh2020mind}, biasing RoIs toward the center; (2) \textit{Noisy activations:} fluctuations in saliency maps may lead to spatially and temporally inconsistent RoIs across frames; and (3) \textit{Variable RoI scale:} appropriate RoI size varies across datasets, action types, and camera views, so a fixed scale may fail to capture all relevant regions. 

We address these challenges as follows. (1) \textit{Center bias removal:} zero-padding in the low-resolution backbone $\phi_l$ is replaced with replicate padding, mitigating such artificial emphasis toward the center of the frames. (2) \textit{Spatio-temporal consistency:} a spatio-temporal Gaussian smoothing filter is applied to $S$, yielding $\tilde{S}$. This reduces noise in the saliency maps and ensures they are spatio-temporally consistent, providing a stable basis for subsequent RoI selection. (3) \textit{Scale adaptivity:} to derive RoIs that flexibly adjust to the saliency spread, we first normalize each frame map $\tilde{S}_l$ such that its values sum up to 1, interpreting $\tilde{S}_l(x, y)$ as a spatial importance probability at each location. For each frame, we select a single RoI $\mathcal{R}_l$, as, for the studied task and datasets, the relevant regions are typically concentrated in a single location, making multiple regions unnecessary (see Supp. \textcolor{blue}{C}). We define $\mathcal{R}_l$ as the smallest rectangular region with a fixed aspect ratio that captures a cumulative importance above a threshold $\tau$, \ie, $\sum_{(x, y)\in \mathcal{R}_l}\tilde{S}_t(x, y) \geq \tau$, while satisfying a minimum region size $(W_r, H_r)$. The threshold $\tau$ controls the tightness of the resulting region and is set empirically. 

\noindent The resulting set of RoIs, $\mathcal{R} = \{\mathcal{R}_l\}_{l=0}^{L-1}$, are cropped from the high-resolution clips $X_h\in \mathbb{R}^{L \times W_h \times H_h \times 3}$ and resized to a fixed size $(W_r, H_r)$ for efficient batched processing, yielding high-resolution RoI clips $X_r \in \mathbb{R}^{L\times W_r \times H_r \times 3}$. This enables \textbf{AdaSpot} to select RoIs that are task-aware, spatially unbiased, consistent across frames, and scale-adaptive, spending high-resolution computation on the most informative regions for precise event localization.

\subsubsection{High-resolution feature extractor}

The high-resolution feature extractor $\phi_h$ operates on the high-resolution RoI clips $X_r$, producing fine-grained representations of the selected regions, \ie, $F_h=\phi_h(X_r) \in \mathbb{R}^{L\times d}$. Its architecture mirrors that of $\phi_l$, but with independent parameters to ensure specialized low- and high-resolution feature representations.

\subsubsection{Temporal modeler}

The temporal modeler integrates complementary information from both the low- and high-resolution branches (\ie, global context from $F_l$ and fine-grained details from $F_h$) while modeling longer-term temporal dependencies.

\mysection{Feature Alignment and Fusion.} Before fusion, each branch's features are projected using lightweight two-layer MLPs with an intermediate ReLU to facilitate distributional alignment, \ie, $F_l' = \phi_l^{\text{proj}}(F_l)$ and $F_h' = \phi_h^{\text{proj}}(F_h)$. The aligned features are then fused via a max-pooling operation $F_f = max(F_l', F_h')\in \mathbb{R}^{L\times d}$, which is both effective and computationally efficient. As detailed in Supp. \textcolor{blue}{C}, more complex fusion mechanisms do not yield significant improvements while increasing computational costs.

\mysection{Temporal modeling.} The fused representations $F_f$ are subsequently processed by a bidirectional GRU layer to capture longer-range temporal dependencies, $F_t=GRU(F_f) \in \mathbb{R}^{L\times 2d}$. We adopt the GRU as our temporal model, as it has demonstrated strong performance in PES~\cite{hong2022spotting, santra2025precise}. 

\subsubsection{Prediction head}

The prediction head produces per-frame class probabilities including a background class. A single linear layer $\phi_{\text{pred}}$ maps $F_t$ to logits $\hat{y} =  \phi_{\text{pred}}(F_t) \in \mathbb{R}^{L \times (C+1)}$, which are converted to probabilities via softmax.

\subsection{Training details}

Following ~\citet{hong2022spotting}, we formulate PES as frame-level classification. At each frame $l$, the model predicts class probabilities $\hat{y}_l$, which are compared against the one-hot ground-truth labels $y_l$ using a weighted cross-entropy loss: $\mathcal{L}_f=\frac{1}{L}\sum_{l=0}^{L-1}CE_w(y_l, \hat{y}_l)$, where $w$ is a scalar weight to balance foreground and background classes.

\mysection{Auxiliar supervision.} Training with only $\mathcal{L}_f$ is, however, unstable (see \cref{sec:component}). To stabilize optimization and encourage both low- and high-resolution branches to learn discriminative and complementary features, we introduce auxiliary supervision at each branch. Specifically, we attach identical temporal modeling and prediction heads --a GRU layer followed by a linear classifier-- to the low- and high-resolution feature streams, $F_l$ and $F_h$, and compute auxiliary weighted cross-entropy losses $\mathcal{L}_l$ and $\mathcal{L}_h$, respectively.

\noindent The overall loss is a weighted combination, $\mathcal{L} = \lambda_f \mathcal{L}_f + \lambda_l \mathcal{L}_l + \lambda_h \mathcal{L}_h$, with $\lambda_f, \lambda_l, \lambda_h$ controlling the contribution of each term. This formulation enforces that (i) the low-resolution branch learns stable, task-relevant features for reliable RoI selection, and (ii) the high-resolution branch captures fine-grained details that naturally complement those from the low-resolution branch. In practice, this one-stage scheme provides robust and stable end-to-end training.

\subsection{Inference}

At inference time, we use clips with 50\% overlapping. Moreover, to reduce the number of candidate events, Soft Non-Maximum Suppression~\cite{bodla2017soft} is applied. Additionally, the auxiliary supervision modules used in the low- and high-resolution branches are discarded.


\begin{table*}[t]
  \caption{Comparison with state-of-the-art methods for PES on Tennis, FineDiving, FineGym, and F3Set datasets. Bold and underlined values indicate the best and second-best results, respectively. The number of parameters (in millions) and GFLOPs for each method are also reported. For AdaSpot, we show results for two variants AdaSpot$^s$ and AdaSpot$^b$ (using RegNetY-200MF and RegNetY-400MF as feature extractors, respectively), reporting the mean over three random seeds along with the standard deviation (\std{std}). $^\dagger$ denotes results obtained by re-running inference with the provided checkpoints using SNMS with a window of two and 50\% overlapping for a fair comparison.}
  \label{tab:sota}
  \centering
  \resizebox{\linewidth}{!}{
  \begin{tabular}{lc>{\columncolor[gray]{0.95}}c>{\columncolor[gray]{0.95}}cc>{\columncolor[gray]{0.95}}c>{\columncolor[gray]{0.95}}cc>{\columncolor[gray]{0.95}}c>{\columncolor[gray]{0.95}}cc>{\columncolor[gray]{0.95}}c>{\columncolor[gray]{0.95}}ccc}
    \toprule
   & \multicolumn{3}{c}{Tennis}  & \multicolumn{3}{c}{FineDiving} & \multicolumn{3}{c}{FineGym} & \multicolumn{3}{c}{F3Set} & \multicolumn{2}{c}{Cost} \\
   \cmidrule(lr){2-4} \cmidrule(lr){5-7} \cmidrule(lr){8-10} \cmidrule(lr){11-13} \cmidrule(lr){14-15}
   
     Model & $\delta=0$f & $1$f & $2$f & $0$f & $1$f & $2$f & $0$f & $1$f & $2$f & $0$f & $1$f & $2$f & P(M) & GFLOPs \\
    \midrule

    E2E-Spot$_{200\text{MF}}$~\cite{hong2022spotting} & 69.78$^\dagger$ & 97.01$^\dagger$ & 97.68$^\dagger$  & 25.00$^\dagger$ & 69.01$^\dagger$ & 86.24$^\dagger$ & 17.50$^\dagger$ & 53.44$^\dagger$ & 63.73$^\dagger$ & -- & -- & -- & 4.49 & 23.13 \\
    E2E-Spot$_{800\text{MF}}$~\cite{hong2022spotting} & 70.04$^\dagger$ & 97.31$^\dagger$ & \underline{97.86}$^\dagger$  & 19.47$^\dagger$ & 65.49$^\dagger$ & 83.83$^\dagger$ & 17.90$^\dagger$ & \textbf{55.05$^\dagger$} & 66.06$^\dagger$ & -- & -- & -- & 12.70 & 84.93 \\

    UGLF~\cite{tran2024unifying} & -- & -- & -- & -- & 70.00 & \textbf{87.70} & -- & 50.20 & \textbf{67.80} & -- & -- & -- & -- & -- \\

    T-DEED$_{200\text{MF}}$~\cite{xarlest} & 56.00$^\dagger$ & 96.91$^\dagger$ & 97.84$^\dagger$ & 21.33$^\dagger$ & 71.07$^\dagger$ & 86.87$^\dagger$ & 17.32$^\dagger$ & 52.99$^\dagger$ & 63.69$^\dagger$ & -- & -- & -- & 16.42 & 21.97 \\
    T-DEED$_{800\text{MF}}$~\cite{xarlest} & 58.43$^\dagger$ & \underline{97.34}$^\dagger$ & \textbf{97.97}$^\dagger$ & 19.63$^\dagger$ & 69.37$^\dagger$ & 85.50$^\dagger$ & \textbf{18.35}$^\dagger$ & 53.97$^\dagger$ & 64.99$^\dagger$ & -- & -- & -- &  64.26 & 86.34 \\

    Santra et al.~\cite{santra2025precise} & 61.01 & 96.21 & 97.75 & -- & -- & -- & 15.24 & 52.31 & \underline{66.57} & -- & -- & -- & 6.46 & 57.84\\
    F$^3$ED~\cite{liu2025f} & -- & -- & -- & -- & -- & -- & -- & -- & -- & 24.79 & 60.71 & 64.79 & -- & -- \\

    \midrule
    \textbf{AdaSpot$^s$} & \underline{73.49}\std{1.2} & 97.28\std{0.1} & 97.76\std{0.1} & \textbf{27.26}\std{1.9} & \underline{71.78}\std{0.9} & \underline{87.66}\std{0.4} & 17.52\std{0.1} & 54.08\std{0.4} & 64.41\std{0.3} & \underline{53.55}\std{1.2} & \underline{67.76}\std{0.8} & \underline{68.41}\std{1.0} & 7.58 & 29.78 \\
    \textbf{AdaSpot$^b$} & \textbf{74.02}\std{1.4} & \textbf{97.36}\std{0.1} & 97.79\std{0.1} & \underline{27.07}\std{1.8} & \textbf{72.00}\std{1.2} & 87.45\std{0.9} & \underline{18.21}\std{0.2} & \underline{54.66}\std{0.2} & 65.24\std{0.2} & \textbf{55.38}\std{0.3} & \textbf{69.37}\std{0.2} & \textbf{69.94}\std{0.2} & 10.63 & 56.78 \\
    \bottomrule
    \end{tabular}
    }
  \vspace{-0.5cm}
\end{table*}

\section{Experiments}

\subsection{Evaluation setup}

\mysection{Datasets.} We evaluate AdaSpot on four datasets. Following~\citet{hong2022spotting}, we use Tennis~\cite{zhang2021vid2player}, FineDiving~\cite{xu2022finediving}, and FineGym~\cite{shao2020finegym} under the PES setting, along with F3Set~\cite{liu2025f}, which targets more fine-grained events. We also evaluate on SoccerNet Ball Action Spotting (SN-BAS)~\cite{soccerNetBallActionSpotting, deliege2021soccernet} under the less strict ES setting, which requires lower temporal precision, to assess AdaSpot's effectiveness in both scenarios. Since SN-BAS has mainly been used in challenge settings~\cite{cioppa2023soccernet, cioppa2024soccernet}, where results often rely on dataset-specific tricks or additional data, we introduce a standardized evaluation protocol (see Supp.~\textcolor{blue}{A}) for fair and reproducible benchmarking. While these datasets focus on sports due to their suitability for PES, AdaSpot is broadly applicable to other domains requiring high temporal precision.

\mysection{Evaluation.} We follow standard practice by training on the training split, using the validation split for early stopping, and reporting results on the test split for all datasets. Performance is measured using mean Average Precision at a temporal tolerance (mAP$@$$\delta$). For Tennis, FineDiving, FineGym, and F3Set, we adopt a strict evaluation protocol with tolerances of $\delta \in \{0, 1, 2\}$ frames. For SN-BAS, we report mAP at temporal tolerances of $\delta \in \{0.5, 1.0\}$ seconds.

\mysection{Implementation details.} For all datasets, videos are first set to a high-resolution format of $796 \times 448$. Following~\cite{hong2022spotting}, Tennis, FineGym, and F3Set are centrally cropped to $(W_h, H_h) = 448 \times 448$. Unless stated otherwise, low-resolution inputs are set to $(W_l, H_l) = \tfrac{1}{2}(W_h, H_h)$. Also following~\cite{hong2022spotting}, for FineDiving, the low-resolution clips are resized to a square format by setting $W_l = H_l$. RoIs are processed at $(W_r, H_r) = (112, 112)$. We report results for two AdaSpot variants --small and big (AdaSpot$^s$ and AdaSpot$^b$)-- which differ in model size and,  hence, computational cost. AdaSpot$^s$ uses RegNetY-200MF and AdaSpot$^b$ uses RegNetY-400MF (both with GSF modules) as feature extractors $\phi_l$ and $\phi_h$. During training, we apply standard data augmentation techniques, including horizontal flipping, gaussian blur, color jitter, affine transformations, and mixup~\cite{zhang2017mixup}. All results are averaged over three runs with different random seeds for robustness. Additional implementation details are provided in Supp.~\textcolor{blue}{B}.

\subsection{Comparison to SOTA}

We compare our method, AdaSpot, with state-of-the-art spotting approaches under both PES and ES settings, focusing on the stricter metrics (mAP$@0$f \textcolor{blue}{for PES} and mAP$@0.5$s \textcolor{blue}{for ES}), while also reporting results on looser metrics for completeness. Results under PES are shown in~\cref{tab:sota}. In \textbf{Tennis} and \textbf{FineDiving} both AdaSpot variants (AdaSpot$^s$ and AdaSpot$^b$) achieve state-of-the-art performance. Notably, the largest gains are on the strictest metric (mAP$@0$f), improving over the best-performing competitor by $+3.98$ on Tennis and $+2.26$ on FineDiving. These results highlight AdaSpot's ability to capture fine-grained temporal cues crucial for precise event localization. In \textbf{FineGym}, AdaSpot also delivers strong results: AdaSpot$^s$ matches or exceeds methods with similar computational cost (\ie, E2E-Spot$_{200\text{MF}}$, T-DEED$_{200\text{MF}}$, and ~\citet{santra2025precise}), while AdaSpot$^b$ outperforms most competitors and achieves performance on par with the best method, T-DEED$_{800\text{MF}}$, using 6x fewer parameters and 1.5x fewer FLOPs. On \textbf{F3Set}, AdaSpot achieves SOTA results on both strict and loose metrics, surpassing F$^3$ED with both variants, showing strong performance on more fine-grained events. In \textbf{SN-BAS under the ES setting}, similar trends are observed (see~\cref{tab:sota2}). AdaSpot$^s$ outperforms all methods with comparable computational cost, and among higher-cost approaches only E2E-Spot$_{800\text{MF}}$ surpasses it. In contrast, AdaSpot$^b$ exceeds E2E-Spot$_{800\text{MF}}$ by $+1.75$ mAP$@0.5$s with 1.66x fewer FLOPs. Overall, AdaSpot consistently achieves strong performance across datasets, offering a superior accuracy-efficiency trade-off relative to prior work. Per-class evaluations in Supp.~\textcolor{blue}{G} show these improvements are consistent across most event categories.

\begin{table}[t]
  \caption{Comparison of AdaSpot with state-of-the-art methods for the ES setting on SN-BAS. Bold and underlined values indicate the best and second-best results. The number of parameters (in millions) and GFLOPs for each method are also reported. Results for E2E-Spot and T-DEED are obtained by integrating their model into our training pipeline. For ~\citet{santra2025precise}, we implement our own version of the ASTRM module following the paper specifications due to the lack of publicly available code. For AdaSpot, we show results for two variants, AdaSpot$^s$ and AdaSpot$^b$, and report the mean over three random seeds with standard deviation (\std{std}).}
  \label{tab:sota2}
  \centering
  \resizebox{\columnwidth}{!}{
  \begin{tabular}{lc>{\columncolor[gray]{0.95}}ccc}
    \toprule
   & \multicolumn{2}{c}{SN-BAS} & \multicolumn{2}{c}{Cost} \\
   \cmidrule(lr){2-3} \cmidrule(lr){4-5}
   
     Model & $\delta$ = $0.5$s & $1$s & P(M) & GFLOPs\\
    \midrule
    E2E-Spot$_{200\text{MF}}$~\cite{hong2022spotting} & 51.46 & 55.11 & 4.49 & 40.78 \\
    E2E-Spot$_{800\text{MF}}$~\cite{hong2022spotting} & \underline{54.49} & \underline{58.65} & 12.70 & 150.02 \\
    T-DEED$_{200\text{MF}}$~\cite{xarlest} & 45.43 & 48.41 & 12.31 & 39.58 \\
    T-DEED$_{800\text{MF}}$~\cite{xarlest} & 49.39 & 53.11 & 46.22 & 151.31\\
    Santra et al.~\cite{santra2025precise} & 51.07 & 55.13 & 6.84 & 82.51\\
    \midrule
    \textbf{AdaSpot$^s$} & 53.12\std{1.4} & 56.82\std{1.9} & 7.58 & 46.18\\
    \textbf{AdaSpot$^b$} & \textbf{56.24}\std{0.3} & \textbf{59.82}\std{0.9} & 10.63 & 90.04 \\
    \bottomrule
    \end{tabular}
    }
  \vspace{-0.5cm}
\end{table}

\subsection{Ablations}

In this section, we conduct ablation studies to validate the design choices of our approach. Specifically, we first analyze the key components of our PES framework, and then compare it with alternative redundancy-aware strategies previously proposed for action recognition. Experiments are conducted on Tennis and SN-BAS. For ablations, we adopt the AdaSpot$^s$ configuration and disable mixup to ensure a more stable component evaluation.

\mysection{Component analysis.}
\label{sec:component}\cref{tab:ablationsTask} summarizes the contribution of each component in AdaSpot. \cref{tab:ablationsTask}(a) compares AdaSpot with its \textbf{single-branch counterparts}. Adding the high-resolution branch to the standard low-resolution pathway improves performance on the stricter tolerances (+2.12 on Tennis and +5.04 on SN-BAS), demonstrating the value of fine-grained spatial details. While using the high-resolution branch alone already surpasses the low-resolution baseline, fusing both provides the best results, proving the effectiveness of combining global context and detailed local cues. In \cref{tab:ablationsTask}(b), we analyze the effect of \textbf{padding type}. Replacing zero-padding with reflect padding improves performance. We attribute this to the center bias introduced by zero-padding~\cite{alsallakh2020mind}, which can lead to biased saliency maps and suboptimal RoI selection (see Supp.~\textcolor{blue}{C}). Reflect padding alleviates this bias, producing more informative RoIs. \cref{tab:ablationsTask}(c) examines the impact of \textbf{spatio-temporal smoothing}. Using raw activations for RoI selection degrades performance (-1.63 on Tennis, -3.03 on SN-BAS), suggesting that noisy saliency produces unstable and inconsistent RoIs. Spatio-temporal smoothing alleviates this, with temporal smoothing contributing most, highlighting the importance of temporally coherent RoIs for effective high-resolution modeling. In \cref{tab:ablationsTask}(d), we analyze \textbf{adaptive RoI scale}. On Tennis, adaptive RoIs outperform a fixed crop size (+0.49 mAP$@0$f), as relevant regions can range from close-up events requiring larger RoIs to distant events where smaller RoIs suffice. In contrast, on SN-BAS, a fixed RoI size performs best, as the distant views of the dataset make the minimum region $(W_r, H_r)$ sufficient to capture the relevant region. Our threshold-based selector unifies both behaviors via a single hyperparameter, allowing the model to adaptively switch between fixed (\ie, $\tau=0$) and adaptive scales depending on dataset characteristics. Finally, \cref{tab:ablationsTask}(e) highlights the importance of \textbf{auxiliary supervision} for stable training. Low-resolution supervision ensures the low-resolution branch learns discriminative features independently, producing reliable saliency maps for RoI selection. High-resolution supervision is equally crucial: without it, early unreliable RoIs can misdirect training, causing the model to largely ignore high-resolution information and perform near the low-resolution-only baseline. Additional analysis on alternative fusion mechanisms, crop sizes, the $\tau$ parameter, multiple regions per frame, and the possibility of reusing the backbone for both branches for improved parameter efficiency are provided in Supp.~\textcolor{blue}{C}.

\begin{table}[t]
  \caption{Ablation study of AdaSpot components on Tennis and SN-BAS, evaluating the impact of single branches, padding types, smoothing methods, fixed versus adaptive RoI scales, and auxiliary supervision. We report the mean over three random seeds along with the standard deviation (\std{std}).}
  \label{tab:ablationsTask}
  \centering
  \resizebox{\columnwidth}{!}{
    \begin{tabular}{llc>{\columncolor[gray]{0.95}}c>{\columncolor[gray]{0.95}}cc>{\columncolor[gray]{0.95}}c}
      \toprule
      & & \multicolumn{3}{c}{Tennis} & \multicolumn{2}{c}{SN-BAS} \\
      \cmidrule(lr){3-5} \cmidrule(lr){6-7}
      \multicolumn{2}{l}{\textbf{Experiment}} & $\delta=0$f & $1$f & $2$f & $\delta=0.5$s & $1$s  \\
      \midrule
      \multicolumn{2}{l}{AdaSpot} & 73.30\std{0.5} & 96.90\std{0.1} & 97.47\std{0.1} & 53.02\std{0.5} & 56.43\std{0.3} \\
      \midrule
      (a) & \textit{Single-branch counterparts} & & & & & \\
      & \quad low-res branch only & 71.18 & 96.73 & 97.42 & 47.98 & 51.69 \\
      & \quad high-res branch only & 71.91 & 96.62 & 97.26 & 52.13 & 55.95 \\
      \midrule
      (b) & \textit{Padding type} & & & & &  \\
      & \quad zero-padding & 72.15 & 96.41 & 96.98 & 51.01 & 54.51 \\
      \midrule
      (c) & \textit{Spatio-temporal smoothing} & & & & &  \\
      & \quad w/o spatial smoothing & 72.63 & 96.79 & 97.37 & 49.43 & 53.04 \\
      & \quad w/o temporal smoothing & 71.86 & 96.56 & 97.17 & 49.29 & 52.67 \\
      & \quad w/o smoothing & 71.67 & 96.87 & 97.43 & 49.99 & 53.84 \\
      \midrule
      (d) & \textit{RoI scale} & & & & &  \\
      & \quad fixed ($\tau=0$) & 72.81 & 96.69 & 97.39 & \textcolor{gray}{53.02} & \textcolor{gray}{56.43} \\
      & \quad adaptive & \textcolor{gray}{73.30} & \textcolor{gray}{96.90} & \textcolor{gray}{97.47} & 51.71 & 55.88 \\
      \midrule
      (e) & \textit{Auxiliary supervision} & & & & &  \\
      & \quad w/o $\mathcal{L}_l$ & 72.81 & 96.91 & 97.49 & 49.71 & 53.27 \\
      & \quad w/o $\mathcal{L}_h$ & 71.18 & 96.40 & 97.12 & 49.48 & 53.12 \\
      & \quad w/o $\mathcal{L}_l$ \& $\mathcal{L}_h$ & 70.49 & 96.25 & 96.96 & 49.22 & 53.17 \\
      \bottomrule
    \end{tabular}
  }
  \vspace{-0.5cm}
\end{table}

\begin{figure}[t]
    \centering
    \begin{minipage}[b]{0.495\columnwidth}
        \centering
        \includegraphics[width=\linewidth]{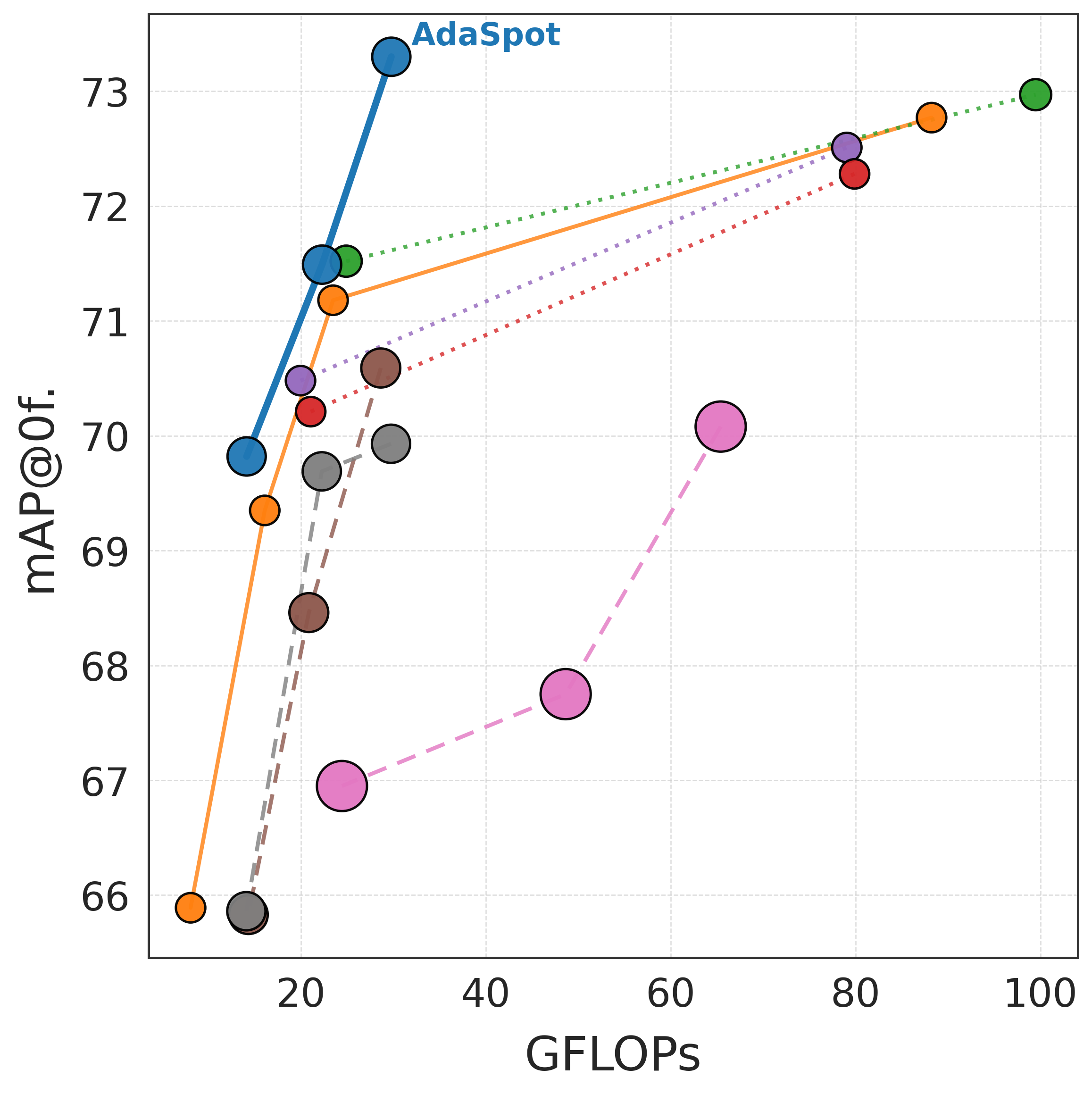}
    \end{minipage}
    \hfill
    \begin{minipage}[b]{0.495\columnwidth}
        \centering
        \includegraphics[width=\linewidth]{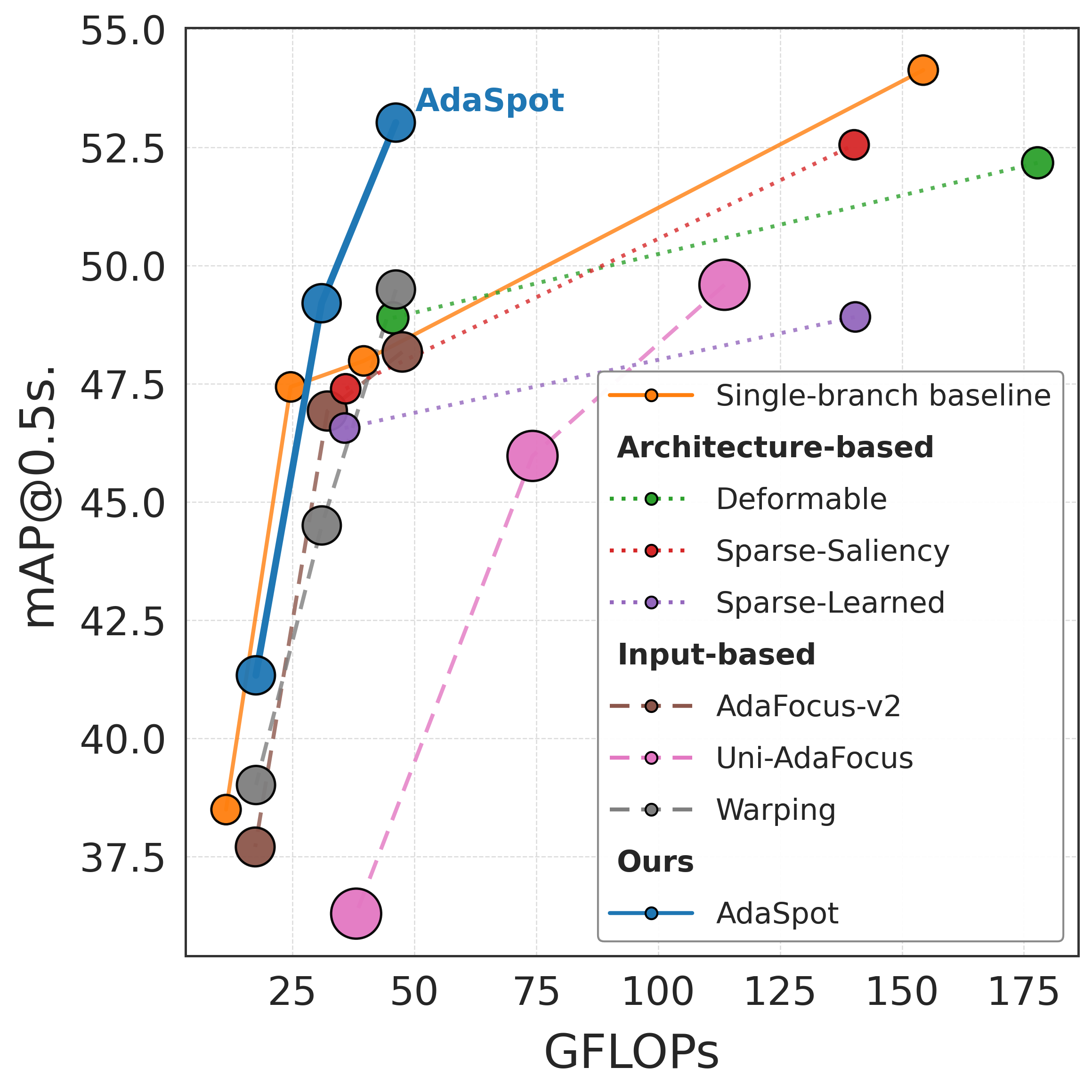}
    \end{minipage}
    \caption{Comparison of AdaSpot, a single-branch baseline, and redundancy-aware alternatives across multiple spatial resolutions on Tennis (left) and SN-BAS (right). Each point corresponds to a model configuration, with GFLOPs on the x-axis, mAP on the y-axis, and point size indicating the number of parameters. Models closer to the upper-left with smaller markers achieve better accuracy-efficiency trade-offs.}
    \label{fig:combined}
    \vspace{-0.5cm}
\end{figure}

\mysection{Comparison with redundancy-aware methods.}
\label{sec:redundancy}
In \cref{fig:combined}, we compare AdaSpot with a single low-resolution branch baseline and representative redundancy-aware approaches for action recognition, evaluated across multiple spatial resolutions. For architecture-based methods, we evaluate deformable~\cite{dai2017deformable} and sparse~\cite{liu2015sparse} convolutions, the latter implemented in two variants: \textit{Sparse-Learned}, where sparsity locations are learned via a gating mechanism, and \textit{Sparse-Saliency}, where they are guided by saliency maps. Input-based methods include learnable pixel-space cropping (AdaFocus-v2~\cite{adafocus-v2}), learnable feature-space cropping with variable-size regions (Uni-AdaFocus~\cite{adafocus-v4}), and saliency-driven frame-warping~\cite{liu2022task} (implementation details in Supp.~\textcolor{blue}{B}). Among \textbf{architecture-based} approaches, efficiency gains are limited. Deformable convolutions slightly increase computational cost relative to dense convolutions and offer only moderate improvements in low-FLOP regimes. Sparse convolutions reduce FLOPs but also degrade accuracy, resulting in a trade-off comparable to the low-resolution baseline. Among \textbf{input-based} approaches, learnable cropping perform poorly when transferred to the PES setting. As analyzed in Supp.~\textcolor{blue}{C}, selected RoIs often fail to cover task-relevant regions, introducing noise during training and reducing the utility of the high-resolution branch. We attribute these limitations of learnable cropping to inherent training instabilities previously observed in action recognition~\cite{adafocus-v4}, which are further exacerbated in the PES setting by weaker supervision signals arising from the highly localized spatio-temporal nature of PES actions. Saliency-based frame warping also does not consistently improve the accuracy-efficiency trade-off. We hypothesize geometric distortions and temporal misalignment introduced during warping hinder spatio-temporal modeling and limit performance gains. Overall, \textbf{AdaSpot} achieves the best accuracy-efficiency trade-off, surpassing the single-branch baseline and alternative redundancy-aware methods with a simple design. For instance, on Tennis, adding our high-resolution branch to the baseline improves mAP$@0$f by $+3.93$, $+2.14$, and $+2.12$ for base resolutions of $112\times 112$, $168\times 168$, and $224\times 224$, respectively, at only a marginal additional computational cost of approximately $+6$ GFLOPs. This improvement exceeds what could be obtained by uniformly increasing the baseline resolution at the same computational cost, highlighting AdaSpot's effectiveness. Similar trends are observed on SN-BAS, with slightly more moderate gains in low-FLOP regimes.

\subsection{Qualitative results}
\label{sec:qualitative}

\begin{figure}[t]
    \centering
    \includegraphics[width=\linewidth]{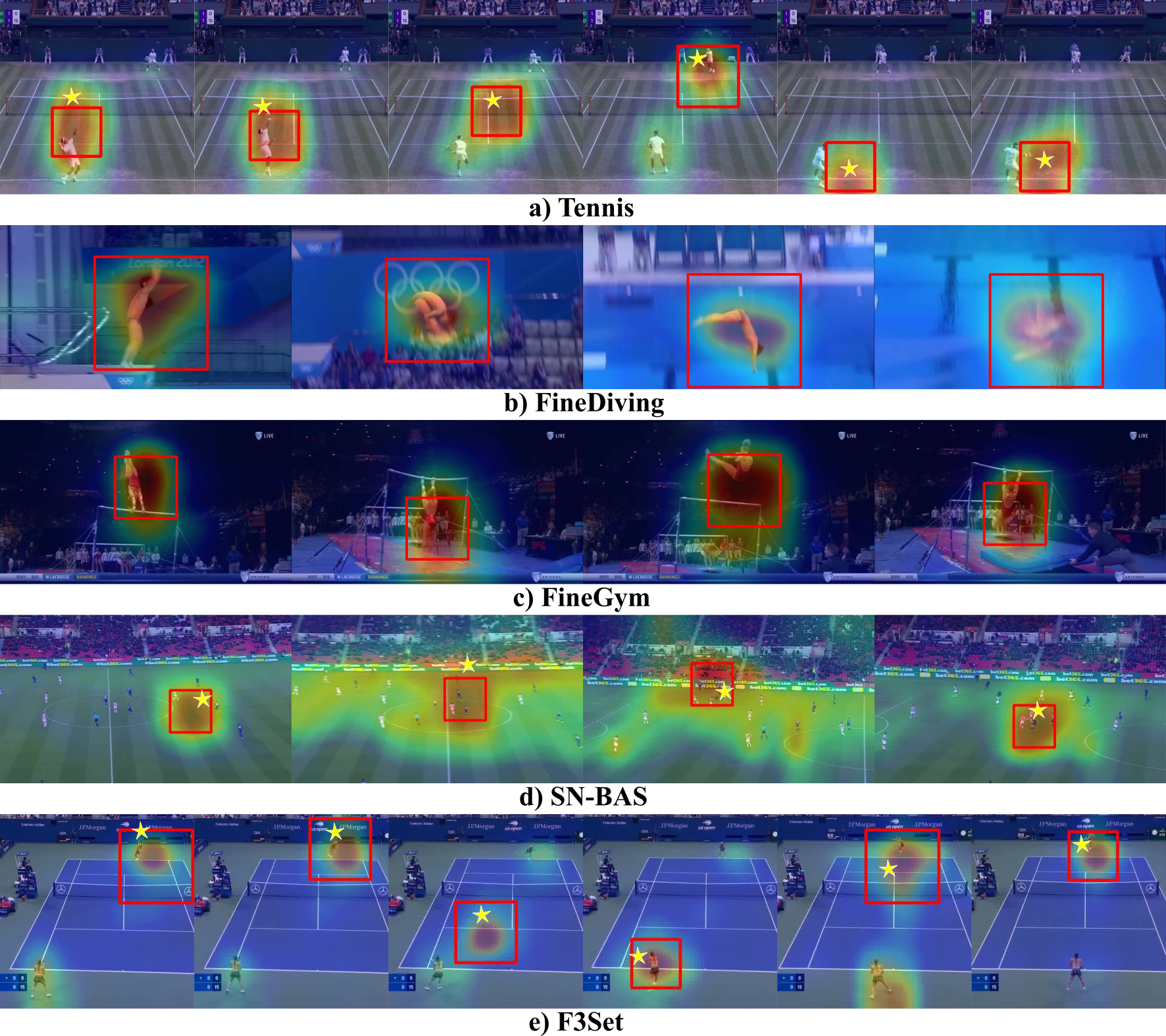}
    \caption{Qualitative visualization of the saliency maps and the corresponding RoIs selected by \textbf{AdaSpot} across all evaluated datasets: Tennis, FineDiving, FineGym, F3Set, and SN-BAS. In FineDiving and FineGym, events revolve around a main athlete, whereas in Tennis, F3Set, and SN-BAS, they revolve around the ball, which is marked with a star for clarity.}
    \label{fig:qualitative}
    \vspace{-0.5cm}
\end{figure}

\Cref{fig:qualitative} shows qualitative examples of the generated saliency maps and the resulting RoIs across datasets. Overall, AdaSpot produces spatially coherent and semantically meaningful RoIs with high temporal consistency. In FineDiving and FineGym, where events revolve around a main athlete, saliency consistently focuses on the athlete, yielding stable RoIs over time. In Tennis and F3Set, where events center on the ball and alternate between two players, more ambiguity is introduced; nevertheless, the model attends to event-relevant regions, with highest saliency on the ball and active player. In the more crowded SN-BAS scenes, where actions also revolve around the ball, AdaSpot effectively tracks the ball, demonstrating robustness in multi-actor scenarios. Occasional uncertainty arises in frames without clear action cues (\eg, \cref{fig:qualitative}(d), frames 2–3), but the model quickly recovers once meaningful actions resume. These qualitative observations align with our quantitative results (\cref{sec:component}): high-resolution processing of selected RoIs consistently improves performance, confirming that the chosen regions capture task-relevant information.

\section{Conclusion}

We presented \textbf{AdaSpot}, a PES framework that explicitly addresses spatial redundancy by allocating high-resolution processing only to the most informative region in each video frame. This approach preserves the fine-grained visual cues crucial for precise localization while maintaining strong computational efficiency. Our experiments demonstrate the importance of capturing these details for accurate PES. They also show that our unsupervised RoI selector identifies semantically meaningful and temporally consistent regions while mitigating the instability issues common in existing learnable-cropping alternatives. 

\mysection{Limitations and future work.} While AdaSpot shows strong performance on current PES benchmarks, its generalization beyond sports remains to be evaluated. In particular, scenarios involving simultaneous actions --where multiple regions may be relevant within a single frame-- require further study to assess AdaSpot's adaptability to multi-RoI selection. Future work could also explore more sophisticated temporal modeling or address temporal redundancy to skip non-informative frames. Although we do not identify any direct harmful applications, AdaSpot could be adapted for privacy-sensitive tasks, motivating future work to consider safeguards to prevent unintended misuse. 

\mysection{Acknowledgements.} This work has been partially supported by the Spanish project PID2022-136436NB-I00, by ICREA under the ICREA Academia programme, and is part of the REPAI project, supported by the Grundfos Foundation. Part of the work was conducted during the first author’s research stay at Game On Technologies. The authors thank Game On Technologies for hosting and supporting the project.

{
    \small
    \bibliographystyle{ieeenat_fullname}
    \bibliography{main}

@String(CVPR= {IEEE Conf. Comput. Vis. Pattern Recog.})

@String(ICPR = {Int. Conf. Pattern Recog.})

@String(TOG= {ACM Trans. Graph.})

@String(ICIP = {IEEE Int. Conf. Image Process.})

@String(CVPR  = {CVPR})

@String(ICPR  = {ICPR})

@String(TOG   = {ACM TOG})

@String(ICIP  = {ICIP})

@article{xia2020survey,
  title={A survey on temporal action localization},
  author={Xia, Huifen and Zhan, Yongzhao},
  journal={IEEE Access},
  volume={8},
  pages={70477--70487},
  year={2020},
  publisher={IEEE}
}

@article{xu2025action,
  title={Action Spotting and Precise Event Detection in Sports: Datasets, Methods, and Challenges},
  author={Xu, Hao and Baniya, Arbind Agrahari and Well, Sam and Bouadjenek, Mohamed Reda and Dazeley, Richard and Aryal, Sunil},
  journal={arXiv preprint arXiv:2505.03991},
  year={2025}
}

@inproceedings{hong2022spotting,
  title={Spotting temporally precise, fine-grained events in video},
  author={Hong, James and Zhang, Haotian and Gharbi, Micha{\"e}l and Fisher, Matthew and Fatahalian, Kayvon},
  booktitle={European Conference on Computer Vision},
  pages={33--51},
  year={2022},
  organization={Springer}
}

@inproceedings{xarles2024t,
  title={T-deed: Temporal-discriminability enhancer encoder-decoder for precise event spotting in sports videos},
  author={Xarles, Artur and Escalera, Sergio and Moeslund, Thomas B and Clap{\'e}s, Albert},
  booktitle={Proceedings of the IEEE/CVF Conference on Computer Vision and Pattern Recognition},
  pages={3410--3419},
  year={2024}
}

@article{xarlest,
  title={T-DEED Revisited: Broader Evaluations and Insights in Precise Event Spotting},
  author={Xarles, Artur and Escalera, Sergio and Moeslund, Thomas B and Clap{\'e}s, Albert},
  year={2024}
}

@inproceedings{santra2025precise,
  title={Precise Event Spotting in Sports Videos: Solving Long-Range Dependency and Class Imbalance},
  author={Santra, Sanchayan and Chudasama, Vishal and Wasnik, Pankaj and Balasubramanian, Vineeth N},
  booktitle={Proceedings of the Computer Vision and Pattern Recognition Conference},
  pages={3163--3172},
  year={2025}
}

@inproceedings{tran2024unifying,
  title={Unifying global and local scene entities modelling for precise action spotting},
  author={Tran, Kim Hoang and Do, Phuc Vuong and Ly, Ngoc Quoc and Le, Ngan},
  booktitle={2024 International Joint Conference on Neural Networks (IJCNN)},
  pages={1--8},
  year={2024},
  organization={IEEE}
}

@inproceedings{sudhakaran2020gate,
  title={Gate-shift networks for video action recognition},
  author={Sudhakaran, Swathikiran and Escalera, Sergio and Lanz, Oswald},
  booktitle={Proceedings of the IEEE/CVF conference on computer vision and pattern recognition},
  pages={1102--1111},
  year={2020}
}

@article{sudhakaran2023gate,
  title={Gate-shift-fuse for video action recognition},
  author={Sudhakaran, Swathikiran and Escalera, Sergio and Lanz, Oswald},
  journal={IEEE Transactions on Pattern Analysis and Machine Intelligence},
  volume={45},
  number={9},
  pages={10913--10928},
  year={2023},
  publisher={IEEE}
}

@inproceedings{radosavovic2020designing,
  title={Designing network design spaces},
  author={Radosavovic, Ilija and Kosaraju, Raj Prateek and Girshick, Ross and He, Kaiming and Doll{\'a}r, Piotr},
  booktitle={Proceedings of the IEEE/CVF conference on computer vision and pattern recognition},
  pages={10428--10436},
  year={2020}
}

@inproceedings{shi2023tridet,
  title={Tridet: Temporal action detection with relative boundary modeling},
  author={Shi, Dingfeng and Zhong, Yujie and Cao, Qiong and Ma, Lin and Li, Jia and Tao, Dacheng},
  booktitle={Proceedings of the IEEE/CVF conference on computer vision and pattern recognition},
  pages={18857--18866},
  year={2023}
}

@inproceedings{zhang2022actionformer,
  title={Actionformer: Localizing moments of actions with transformers},
  author={Zhang, Chen-Lin and Wu, Jianxin and Li, Yin},
  booktitle={European Conference on Computer Vision},
  pages={492--510},
  year={2022},
  organization={Springer}
}

@inproceedings{soares2022temporally,
  title={Temporally precise action spotting in soccer videos using dense detection anchors},
  author={Soares, Joao VB and Shah, Avijit and Biswas, Topojoy},
  booktitle={2022 IEEE International Conference on Image Processing (ICIP)},
  pages={2796--2800},
  year={2022},
  organization={IEEE}
}

@article{zhou2021feature,
  title={Feature combination meets attention: Baidu soccer embeddings and transformer based temporal detection},
  author={Zhou, Xin and Kang, Le and Cheng, Zhiyu and He, Bo and Xin, Jingyu},
  journal={arXiv preprint arXiv:2106.14447},
  year={2021}
}

@inproceedings{xarles2023astra,
  title={Astra: An action spotting transformer for soccer videos},
  author={Xarles, Artur and Escalera, Sergio and Moeslund, Thomas B and Clap{\'e}s, Albert},
  booktitle={Proceedings of the 6th International Workshop on Multimedia Content Analysis in Sports},
  pages={93--102},
  year={2023}
}

@inproceedings{denize2024comedian,
  title={COMEDIAN: Self-supervised learning and knowledge distillation for action spotting using transformers},
  author={Denize, Julien and Liashuha, Mykola and Rabarisoa, Jaonary and Orcesi, Astrid and H{\'e}rault, Romain},
  booktitle={Proceedings of the IEEE/CVF Winter Conference on applications of computer vision},
  pages={530--540},
  year={2024}
}

@inproceedings{dai2017deformable,
  title={Deformable convolutional networks},
  author={Dai, Jifeng and Qi, Haozhi and Xiong, Yuwen and Li, Yi and Zhang, Guodong and Hu, Han and Wei, Yichen},
  booktitle={Proceedings of the IEEE international conference on computer vision},
  pages={764--773},
  year={2017}
}

@inproceedings{liu2015sparse,
  title={Sparse convolutional neural networks},
  author={Liu, Baoyuan and Wang, Min and Foroosh, Hassan and Tappen, Marshall and Pensky, Marianna},
  booktitle={Proceedings of the IEEE conference on computer vision and pattern recognition},
  pages={806--814},
  year={2015}
}

@article{child2019generating,
  title={Generating long sequences with sparse transformers},
  author={Child, Rewon and Gray, Scott and Radford, Alec and Sutskever, Ilya},
  journal={arXiv preprint arXiv:1904.10509},
  year={2019}
}

@inproceedings{meng2020ar,
  title={Ar-net: Adaptive frame resolution for efficient action recognition},
  author={Meng, Yue and Lin, Chung-Ching and Panda, Rameswar and Sattigeri, Prasanna and Karlinsky, Leonid and Oliva, Aude and Saenko, Kate and Feris, Rogerio},
  booktitle={European conference on computer vision},
  pages={86--104},
  year={2020},
  organization={Springer}
}

@inproceedings{lin2022ocsampler,
  title={Ocsampler: Compressing videos to one clip with single-step sampling},
  author={Lin, Jintao and Duan, Haodong and Chen, Kai and Lin, Dahua and Wang, Limin},
  booktitle={Proceedings of the IEEE/CVF conference on computer vision and pattern recognition},
  pages={13894--13903},
  year={2022}
}

@inproceedings{ghodrati2021frameexit,
  title={Frameexit: Conditional early exiting for efficient video recognition},
  author={Ghodrati, Amir and Bejnordi, Babak Ehteshami and Habibian, Amirhossein},
  booktitle={Proceedings of the IEEE/CVF Conference on Computer Vision and Pattern Recognition},
  pages={15608--15618},
  year={2021}
}

@inproceedings{karpathy2014large,
  title={Large-scale video classification with convolutional neural networks},
  author={Karpathy, Andrej and Toderici, George and Shetty, Sanketh and Leung, Thomas and Sukthankar, Rahul and Fei-Fei, Li},
  booktitle={Proceedings of the IEEE conference on Computer Vision and Pattern Recognition},
  pages={1725--1732},
  year={2014}
}

@inproceedings{adafocus-v1,
  title={Adaptive focus for efficient video recognition},
  author={Wang, Yulin and Chen, Zhaoxi and Jiang, Haojun and Song, Shiji and Han, Yizeng and Huang, Gao},
  booktitle={proceedings of the IEEE/CVF international conference on computer vision},
  pages={16249--16258},
  year={2021}
}

@inproceedings{adafocus-v2,
  title={Adafocus v2: End-to-end training of spatial dynamic networks for video recognition},
  author={Wang, Yulin and Yue, Yang and Lin, Yuanze and Jiang, Haojun and Lai, Zihang and Kulikov, Victor and Orlov, Nikita and Shi, Humphrey and Huang, Gao},
  booktitle={2022 IEEE/CVF Conference on Computer Vision and Pattern Recognition (CVPR)},
  pages={20030--20040},
  year={2022},
  organization={IEEE}
}

@inproceedings{jana2021unsupervised,
  title={Unsupervised action localization crop in video retargeting for 3D ConvNets},
  author={Jana, Prithwish and Bhaumik, Swarnabja and Mohanta, Partha Pratim},
  booktitle={TENCON 2021-2021 IEEE Region 10 Conference (TENCON)},
  pages={670--675},
  year={2021},
  organization={IEEE}
}

@inproceedings{adafocus-v3,
  title={Adafocusv3: On unified spatial-temporal dynamic video recognition},
  author={Wang, Yulin and Yue, Yang and Xu, Xinhong and Hassani, Ali and Kulikov, Victor and Orlov, Nikita and Song, Shiji and Shi, Humphrey and Huang, Gao},
  booktitle={European Conference on Computer Vision},
  pages={226--243},
  year={2022},
  organization={Springer}
}

@article{adafocus-v4,
  title={Uni-adafocus: spatial-temporal dynamic computation for video recognition},
  author={Wang, Yulin and Zhang, Haoji and Yue, Yang and Song, Shiji and Deng, Chao and Feng, Junlan and Huang, Gao},
  journal={IEEE Transactions on Pattern Analysis and Machine Intelligence},
  year={2024},
  publisher={IEEE}
}

@article{zheng2023dynamic,
  title={Dynamic spatial focus for efficient compressed video action recognition},
  author={Zheng, Ziwei and Yang, Le and Wang, Yulin and Zhang, Miao and He, Lijun and Huang, Gao and Li, Fan},
  journal={IEEE Transactions on Circuits and Systems for Video Technology},
  volume={34},
  number={2},
  pages={695--708},
  year={2023},
  publisher={IEEE}
}

@inproceedings{liu2022task,
  title={Task-adaptive spatial-temporal video sampler for few-shot action recognition},
  author={Liu, Huabin and Lv, Weixian and See, John and Lin, Weiyao},
  booktitle={Proceedings of the 30th ACM International Conference on Multimedia},
  pages={6230--6240},
  year={2022}
}

@inproceedings{xia2022vision,
  title={Vision transformer with deformable attention},
  author={Xia, Zhuofan and Pan, Xuran and Song, Shiji and Li, Li Erran and Huang, Gao},
  booktitle={Proceedings of the IEEE/CVF conference on computer vision and pattern recognition},
  pages={4794--4803},
  year={2022}
}

@inproceedings{kapidis2019egocentric,
  title={Egocentric hand track and object-based human action recognition},
  author={Kapidis, Georgios and Poppe, Ronald and Van Dam, Elsbeth and Noldus, Lucas and Veltkamp, Remco},
  booktitle={2019 IEEE SmartWorld, Ubiquitous Intelligence \& Computing, Advanced \& Trusted Computing, Scalable Computing \& Communications, Cloud \& Big Data Computing, Internet of People and Smart City Innovation (SmartWorld/SCALCOM/UIC/ATC/CBDCom/IOP/SCI)},
  pages={922--929},
  year={2019},
  organization={IEEE}
}

@inproceedings{zhou2016learning,
  title={Learning deep features for discriminative localization},
  author={Zhou, Bolei and Khosla, Aditya and Lapedriza, Agata and Oliva, Aude and Torralba, Antonio},
  booktitle={Proceedings of the IEEE conference on computer vision and pattern recognition},
  pages={2921--2929},
  year={2016}
}

@inproceedings{bodla2017soft,
  title={Soft-NMS--improving object detection with one line of code},
  author={Bodla, Navaneeth and Singh, Bharat and Chellappa, Rama and Davis, Larry S},
  booktitle={Proceedings of the IEEE international conference on computer vision},
  pages={5561--5569},
  year={2017}
}

@inproceedings{xu2022finediving,
  title={Finediving: A fine-grained dataset for procedure-aware action quality assessment},
  author={Xu, Jinglin and Rao, Yongming and Yu, Xumin and Chen, Guangyi and Zhou, Jie and Lu, Jiwen},
  booktitle={Proceedings of the IEEE/CVF conference on computer vision and pattern recognition},
  pages={2949--2958},
  year={2022}
}

@article{zhang2021vid2player,
  title={Vid2player: Controllable video sprites that behave and appear like professional tennis players},
  author={Zhang, Haotian and Sciutto, Cristobal and Agrawala, Maneesh and Fatahalian, Kayvon},
  journal={ACM Transactions on Graphics (TOG)},
  volume={40},
  number={3},
  pages={1--16},
  year={2021},
  publisher={ACM New York, NY}
}

@inproceedings{shao2020finegym,
  title={Finegym: A hierarchical video dataset for fine-grained action understanding},
  author={Shao, Dian and Zhao, Yue and Dai, Bo and Lin, Dahua},
  booktitle={Proceedings of the IEEE/CVF conference on computer vision and pattern recognition},
  pages={2616--2625},
  year={2020}
}

@misc{soccerNetBallActionSpotting,
  author       = {SoccerNet},
  title        = {SoccerNet Ball Action Spotting},
  howpublished = {\url{https://www.soccer-net.org/tasks/ball-action-spotting}},
  year         = {2023},
  note         = {Online; accessed 2025-13-10},
}

@article{cioppa2023soccernet,
  title={SoccerNet 2023 challenges results},
  author={Cioppa, Anthony and Giancola, Silvio and Somers, Vladimir and Magera, Floriane and Zhou, Xin and Mkhallati, Hassan and Deli{\`e}ge, Adrien and Held, Jan and Hinojosa, Carlos and Mansourian, Amir M and others},
  journal={Sports Engineering},
  volume={27},
  number={2},
  pages={24},
  year={2024},
  publisher={Springer}
}

@article{cioppa2024soccernet,
  title         = {SoccerNet 2024 Challenges Results},
  author        = {Cioppa, A. and Giancola, S. and Somers, V. and Joos, V. and Magera, F. and Held, J. and Ghasemzadeh, S. A. and Zhou, X. and Seweryn, K. and Kowalczyk, M. and Mr{\'o}z, Z. and Lukasik, S. and Halo{\'n}, M. and Mkhallati, H. and Deli\`ege, A. and Hinojosa, C. and Sanchez, K. and Mansourian, A. M. and Miralles, P. and Barnich, O. and De Vleeschouwer, C. and Alahi, A. and Ghanem, B. and Van Droogenbroeck, M. and Gorski, A. and Clap{\'e}s, A. and Boiarov, A. and Afanasiev, A. and Xarles, A. and Scott, A. and Lim, B. and Yeung, C. and Gonzalez, C. and R{\"u}fenacht, D. and Pacilio, E. and Deuser, F. and Altawijri, F. S. and Cach{\'o}n, F. and Kim, H. and Wang, H. and Choe, H. and Kim, H. J. and Kim, I.-M. and Kang, J.-M. and Tursunboev, J. and Yang, J. and Hong, J. and Lee, J. and Zhang, J. and Lee, J. and Zhang, K. and Habel, K. and Jiao, L. and Li, L. and Guti{\'e}rrez-P{\'e}rez, M. and Ortega, M. and Li, M. and Lopatto, M. and Kasatkin, N. and Nemtsev, N. and Oswald, N. and Udin, O. and Kononov, P. and Geng, P. and Alotaibi, S. G. and Kim, S. and Ulasen, S. and Escalera, S. and Zhang, S. and Yang, S. and Moon, S. and Moeslund, T. B. and Shandyba, V. and Golovkin, V. and Dai, W. and Chung, W. and Liu, X. and Zhu, Y. and Kim, Y. and Li, Y. and Yang, Y. and Xiao, Y. and Cheng, Z. and Li, Z.},
  journal       = {arXiv preprint arXiv:2409.10587},
  year          = {2024},
  doi           = {10.48550/arXiv.2409.10587},
  url           = {https://doi.org/10.48550/arXiv.2409.10587}
}

@article{loshchilov2017decoupled,
  title={Decoupled weight decay regularization},
  author={Loshchilov, Ilya and Hutter, Frank},
  journal={arXiv preprint arXiv:1711.05101},
  year={2017}
}

@inproceedings{wu2019multi,
  title={Multi-agent reinforcement learning based frame sampling for effective untrimmed video recognition},
  author={Wu, Wenhao and He, Dongliang and Tan, Xiao and Chen, Shifeng and Wen, Shilei},
  booktitle={Proceedings of the IEEE/CVF International Conference on Computer Vision},
  pages={6222--6231},
  year={2019}
}

@article{wu2020dynamic,
  title={A dynamic frame selection framework for fast video recognition},
  author={Wu, Zuxuan and Li, Hengduo and Xiong, Caiming and Jiang, Yu-Gang and Davis, Larry S},
  journal={IEEE Transactions on Pattern Analysis and Machine Intelligence},
  volume={44},
  number={4},
  pages={1699--1711},
  year={2020},
  publisher={IEEE}
}

@inproceedings{korbar2019scsampler,
  title={Scsampler: Sampling salient clips from video for efficient action recognition},
  author={Korbar, Bruno and Tran, Du and Torresani, Lorenzo},
  booktitle={Proceedings of the IEEE/CVF International Conference on Computer Vision},
  pages={6232--6242},
  year={2019}
}

@inproceedings{xia2022nsnet,
  title={Nsnet: Non-saliency suppression sampler for efficient video recognition},
  author={Xia, Boyang and Wu, Wenhao and Wang, Haoran and Su, Rui and He, Dongliang and Yang, Haosen and Fan, Xiaoran and Ouyang, Wanli},
  booktitle={European Conference on Computer Vision},
  pages={705--723},
  year={2022},
  organization={Springer}
}

@article{alsallakh2020mind,
  title={Mind the Pad--CNNs Can Develop Blind Spots},
  author={Alsallakh, Bilal and Kokhlikyan, Narine and Miglani, Vivek and Yuan, Jun and Reblitz-Richardson, Orion},
  journal={arXiv preprint arXiv:2010.02178},
  year={2020}
}

@article{kong2022human,
  title={Human action recognition and prediction: A survey},
  author={Kong, Yu and Fu, Yun},
  journal={International Journal of Computer Vision},
  volume={130},
  number={5},
  pages={1366--1401},
  year={2022},
  publisher={Springer}
}

@article{han2021dynamic,
  title={Dynamic neural networks: A survey},
  author={Han, Yizeng and Huang, Gao and Song, Shiji and Yang, Le and Wang, Honghui and Wang, Yulin},
  journal={IEEE transactions on pattern analysis and machine intelligence},
  volume={44},
  number={11},
  pages={7436--7456},
  year={2021},
  publisher={IEEE}
}

@article{zhang2017mixup,
  title={mixup: Beyond empirical risk minimization},
  author={Zhang, Hongyi and Cisse, Moustapha and Dauphin, Yann N and Lopez-Paz, David},
  journal={arXiv preprint arXiv:1710.09412},
  year={2017}
}

@article{giancola2024deep,
  title={Deep learning for action spotting in association football videos},
  author={Giancola, Silvio and Cioppa, Anthony and Ghanem, Bernard and Van Droogenbroeck, Marc},
  journal={arXiv preprint arXiv:2410.01304},
  year={2024}
}

@article{rafiq2020scene,
  title={Scene classification for sports video summarization using transfer learning},
  author={Rafiq, Muhammad and Rafiq, Ghazala and Agyeman, Rockson and Choi, Gyu Sang and Jin, Seong-Il},
  journal={Sensors},
  volume={20},
  number={6},
  pages={1702},
  year={2020},
  publisher={MDPI}
}

@article{naik2022comprehensive,
  title={A comprehensive review of computer vision in sports: Open issues, future trends and research directions},
  author={Naik, Banoth Thulasya and Hashmi, Mohammad Farukh and Bokde, Neeraj Dhanraj},
  journal={Applied Sciences},
  volume={12},
  number={9},
  pages={4429},
  year={2022},
  publisher={MDPI}
}

@article{sighencea2021review,
  title={A review of deep learning-based methods for pedestrian trajectory prediction},
  author={Sighencea, Bogdan Ilie and Stanciu, Rareș Ion and C{\u{a}}leanu, C{\u{a}}t{\u{a}}lin Daniel},
  journal={Sensors},
  volume={21},
  number={22},
  pages={7543},
  year={2021},
  publisher={MDPI}
}

@inproceedings{zhdanova2020human,
  title={Human activity recognition for efficient human-robot collaboration},
  author={Zhdanova, M and Voronin, V and Semenishchev, E and Ilyukhin, Yu and Zelensky, A},
  booktitle={Artificial Intelligence and Machine Learning in Defense Applications II},
  volume={11543},
  pages={94--104},
  year={2020},
  organization={SPIE}
}

@phdthesis{keshinro2023human,
  title={Human Activity Recognition Using Deep Learning Methods for Human-Robot Interaction},
  author={Keshinro, Babatunde Ibrahim},
  year={2023},
  school={North Carolina Agricultural and Technical State University}
}

@inproceedings{deliege2021soccernet,
  title={Soccernet-v2: A dataset and benchmarks for holistic understanding of broadcast soccer videos},
  author={Deliege, Adrien and Cioppa, Anthony and Giancola, Silvio and Seikavandi, Meisam J and Dueholm, Jacob V and Nasrollahi, Kamal and Ghanem, Bernard and Moeslund, Thomas B and Van Droogenbroeck, Marc},
  booktitle={Proceedings of the IEEE/CVF conference on computer vision and pattern recognition},
  pages={4508--4519},
  year={2021}
}

@inproceedings{dalal2025action,
  title={Action Anticipation from SoccerNet Football Video Broadcasts},
  author={Dalal, Mohamad and Xarles, Artur and Cioppa, Anthony and Giancola, Silvio and Van Droogenbroeck, Marc and Ghanem, Bernard and Clap{\'e}s, Albert and Escalera, Sergio and Moeslund, Thomas B},
  booktitle={Proceedings of the Computer Vision and Pattern Recognition Conference},
  pages={6080--6091},
  year={2025}
}

@article{paszke2019pytorch,
  title={Pytorch: An imperative style, high-performance deep learning library},
  author={Paszke, Adam and Gross, Sam and Massa, Francisco and Lerer, Adam and Bradbury, James and Chanan, Gregory and Killeen, Trevor and Lin, Zeming and Gimelshein, Natalia and Antiga, Luca and others},
  journal={Advances in neural information processing systems},
  volume={32},
  year={2019}
}

@article{bengio2013estimating,
  title={Estimating or propagating gradients through stochastic neurons for conditional computation},
  author={Bengio, Yoshua and L{\'e}onard, Nicholas and Courville, Aaron},
  journal={arXiv preprint arXiv:1308.3432},
  year={2013}
}

@inproceedings{neubeck2006efficient,
  title={Efficient non-maximum suppression},
  author={Neubeck, Alexander and Van Gool, Luc},
  booktitle={18th international conference on pattern recognition (ICPR'06)},
  volume={3},
  pages={850--855},
  year={2006},
  organization={IEEE}
}

@article{liu2025f,
  title={F$^3$ Set: Towards Analyzing Fast, Frequent, and Fine-grained Events from Videos},
  author={Liu, Zhaoyu and Jiang, Kan and Ma, Murong and Hou, Zhe and Lin, Yun and Dong, Jin Song},
  journal={arXiv preprint arXiv:2504.08222},
  year={2025}
}

@article{liu2025few,
  title={Few-Shot Precise Event Spotting via Unified Multi-Entity Graph and Distillation},
  author={Liu, Zhaoyu and Jiang, Kan and Ma, Murong and Hou, Zhe and Lin, Yun and Dong, Jin Song},
  journal={arXiv preprint arXiv:2511.14186},
  year={2025}
}

@inproceedings{mcnally2019golfdb,
  title={Golfdb: A video database for golf swing sequencing},
  author={McNally, William and Vats, Kanav and Pinto, Tyler and Dulhanty, Chris and McPhee, John and Wong, Alexander},
  booktitle={Proceedings of the IEEE/CVF conference on computer vision and pattern recognition workshops},
  pages={0--0},
  year={2019}
}

@inproceedings{voeikov2020ttnet,
  title={TTNet: Real-time temporal and spatial video analysis of table tennis},
  author={Voeikov, Roman and Falaleev, Nikolay and Baikulov, Ruslan},
  booktitle={Proceedings of the IEEE/CVF conference on computer vision and pattern recognition workshops},
  pages={884--885},
  year={2020}
}
}

\clearpage

\appendix

\begin{center}
    \large \textbf{Supplementary Material}
\end{center}

In this supplementary material, we provide additional details and analyses complementing the main paper. We first describe the datasets and evaluation protocols in more depth (\cref{sec:datasets_evaluation}). Next, we expand on the implementation details of AdaSpot and the state-of-the-art methods (\cref{sec:implementation}). \cref{sec:ablation} presents extended ablation studies, while \cref{sec:efficiency} and \cref{sec:randomness} provide further discussions on efficiency and randomness analyses, respectively. In \cref{sec:postprocessing}, we investigate the sensitivity of PES methods to the choice of post-processing. Finally, \cref{sec:additional_results} reports additional per-class and qualitative results for AdaSpot.

\section{Data and evaluation protocols description}
\label{sec:datasets_evaluation}

In this section, we first provide additional details about the datasets, followed by further clarification of the evaluation protocols.

\subsection{Datasets description}

We evaluated AdaSpot on five datasets: Tennis~\cite{zhang2021vid2player}, FineDiving~\cite{xu2022finediving}, FineGym~\cite{shao2020finegym}, and F3Set~\cite{liu2025f} under the Precise Event Spotting (PES) setting, as well as SoccerNet Ball Action Spotting (SN-BAS)~\cite{soccerNetBallActionSpotting, deliege2021soccernet} under the less strict Event Spotting (ES) setting, which requires lower temporal precision. In the following, we provide additional details for each dataset.

\mysection{Tennis.} The Tennis dataset, originally introduced in~\citet{zhang2021vid2player} and later extended by~\citet{hong2022spotting} consists of 3\,345 video clips, each corresponding to a single tennis point, extracted from 28 matches. The videos have frame rates ranging from 25 to 30 frames per second. In total, the dataset contains 33\,791 precisely annotated events across six classes --``serve'', ``swing'', and ``ball bounce'', each distinguished between near- and far-court--, with the class-wise distribution provided in~\cref{tab:tennis}. All annotated events are therefore ball-centric, indicating that the regions of interest in this dataset are predominantly around the ball's spatial location.

\mysection{FineDiving.} The FineDiving dataset, introduced by~\citet{xu2022finediving}, comprises 3\,000 diving clips recorded at 25 frames per second. In total, it contains 7\,010 events corresponding to transitions into somersaults, categorized into four classes --``pike'', ``tuck'', ``twist'', and ``entry''--, with per-class frequencies provided in~\cref{tab:finediving}. All annotated events involve a single primary athlete, so the regions of interest are naturally centered on that athlete. 

\mysection{FineGym.} The FineGym dataset, introduced by~\citet{shao2020finegym}, comprises 5\,374 untrimmed videos of gymnastics performances, originally recorded at frame rates between 25 and 60 frames per second. Following~\citet{hong2022spotting}, the videos' frame rates are standardized to 25-30 fps. In total, the dataset contains 80\,166 events corresponding to the start and end of various gymnastics actions --such as ``floor exercise turns'', ``uneven bars dismounts'', and ``balance beam turns''--, spanning 32 event classes, with per-class frequencies summarized in~\cref{tab:finegym}. As these events are athlete-centric, the regions of interest are predominantly focused on the main athlete. 

\mysection{F3Set.} The F3Set dataset, introduced by~\citet{liu2025f} contains 11\,584 video clips, each corresponding to a tennis point, extracted from 114 matches featuring 75 players. Videos have frame rates between 25 and 30 fps. In total, the dataset includes 42\,846 precisely annotated events across 365 classes, each representing a combination of categories such as the player hitting the ball, court location, body side, shot type, shot direction, shot technique, player movement, and shot outcome. While F3Set is similar to Tennis, it features far more fine-grained events. We omit per-class statistics and analyses due to the large number of classes. As in Tennis, all events are ball-centric.

\mysection{SN-BAS.} The SN-BAS dataset~\cite{soccerNetBallActionSpotting, deliege2021soccernet} consists of untrimmed videos from seven English Football League matches recorded at 25 fps. In total, the dataset contains 12\,422 annotated ball-related events. The event classes include: ``pass'', ``drive'', ``header'', ``high pass'', ``out'', ``cross'', ``throw-in'', ``shot'', ``ball-player block'', ``player successful tackle'', ``free-kick'', and ``goal'', with per-class frequencies listed in~\cref{tab:snbas}. As the events are ball-centric, the relevant regions of interest are naturally centered around the ball. 

\subsection{Evaluation protocols}
\label{sec:evaluationprotocol}

For the Tennis, FineDiving, FineGym, and F3Set datasets under the PES setting, we follow the evaluation protocol proposed by~\citet{hong2022spotting}, using the same training, validation, and test splits. The task is evaluated using mean Average Precision at a given temporal tolerance $\delta$, denoted as $mAP@\delta$. For these datasets, we report results using temporal tolerances of $\delta \in \{0, 1, 2\}$ frames.

For SN-BAS, which has primarily been used for challenge purposes~\cite{cioppa2023soccernet, cioppa2024soccernet}, many existing methods rely on dataset-specific tricks, alternative data splits, or external data sources. To ensure fair and reproducible benchmarking, we introduce a standardized evaluation protocol. We adopt the original data splits, training on the four-game training set, using the one-game validation for early stopping, and reporting results on the two-game test set, while discarding the two-game challenge set with hidden ground-truth. Following~\cite{dalal2025action}, we exclude the ``free-kick'' and ``goal'' event classes due to their extremely low frequency --six and two examples, respectively, in the test split-- which makes the metric highly sensitive to single correct or incorrect predictions. To maintain a more stable and meaningful evaluation, we remove these classes from our analysis. For this dataset, under the less strict ES setting, the task is evaluated using mean Average Precision with temporal tolerances of $\delta \in \{0.5, 1\}$ seconds.

\section{Implementation details}
\label{sec:implementation}

To ensure reproducibility, in this section we provide implementation details for AdaSpot, as well as those for the state-of-the-art models used in our comparisons. We also describe the adaptations required to apply redundancy-aware methods to the PES setting.

\subsection{AdaSpot}
\label{sec:adaspot}

In addition to the implementation details provided in Sec.~\textcolor{blue}{4.1}, we train AdaSpot on clips of $L=100$ frames with a batch size of $4$. For the two model variants, AdaSpot$^s$, and AdaSpot$^b$, which use different feature-extractor sizes, the hidden dimensions are set to $d=368$ and $d=608$, respectively. The RoI selector uses an upsampling factor of $k=8$, and the threshold parameter $\tau$ is empirically tuned for each dataset. To mitigate class imbalance, the cross-entropy-loss assigns a weight of $w=5$ to positive classes, and the loss coefficients are set to $\lambda_f=\lambda_l=\lambda_h=\tfrac{1}{3}$. For F3Set, we use per-event-category prediction heads to handle multiple categories, with each event class probability computed as the product of its category probabilities. Each epoch consists of $5\,000$ randomly sampled clips. We train the models for $25$ epochs on FineDiving, $50$ epochs on Tennis and SN-BAS, and $100$ epochs on the larger FineGym and F3Set datasets. Optimization is performed with AdamW~\cite{loshchilov2017decoupled}, using a base learning rate of $8\mathrm{e}{-4}$, five warm-up epochs, and cosine learning-rate decay. Soft Non-Maximum Suppression uses a window size of $2$ frames for PES and $12$ for ES. The method is implemented in PyTorch~\cite{paszke2019pytorch}, and all models are trained on a single NVIDIA RTX 6000 Ada Generation GPU.

\subsection{State-of-the-art models}

\mysection{PES setting.} In the PES setting, we compare AdaSpot against several state-of-the-art methods. Specifically, we include \underline{E2E-Spot~\cite{hong2022spotting}} in two variants --E2E-Spot$_{200\text{MF}}$ and E2E-Spot$_{800\text{MF}}$-- which use RegNetY-200MF and RegNetY-800MF as feature extractors, respectively. We also include \underline{UGLF~\cite{tran2024unifying}} and \underline{T-DEED~\cite{xarles2024t}}, the latter likewise evaluated in two configurations, T-DEED$_{200\text{MF}}$ and T-DEED$_{800\text{MF}}$, based on the same RegNetY backbones. In addition, we report results for \underline{~\citet{santra2025precise}}, and F$^3$ED for F3Set. We exclude alternative approaches considered in~\citet{hong2022spotting}, such as two-stage methods with pre-extracted features, due to their lower performance, focusing the comparison on end-to-end models that achieve high performance. As discussed in~\cref{sec:postprocessing}, the choice of postprocessing technique can notably affect the evaluation metrics. To ensure a fair comparison, we report results for all methods using the same postprocessing procedure specified in~\cref{sec:ablation}. For E2E-Spot and T-DEED, which originally report results with different postprocessing strategies, we run inference using their publicly available checkpoints and update the postprocessing accordingly. \citet{santra2025precise} already reports results using Soft-NMS with a window of $2$. For UGLF, no public checkpoints are available; thus, we report the results as provided in their original paper, which uses a different postprocessing setup. Finally, for F$^3$ED, we run inference using the publicly available checkpoints and modify the code to compute the mAP metrics not included in the original implementation. We additionally compare AdaSpot and F$^3$ED under their native evaluation metrics in~\cref{sec:additional_results}.

\mysection{ES setting.} In the ES setting, we compare AdaSpot against \underline{E2E-Spot} (in two variants: E2E-Spot$_{200\text{MF}}$ and E2E-Spot$_{800\text{MF}}$), \underline{T-DEED} (T-DEED$_{200\text{MF}}$ and T-DEED$_{800\text{MF}}$), and \underline{~\citet{santra2025precise}}. Since none of these methods provide pre-trained models on SN-BAS under the evaluation protocol specified in~\cref{sec:evaluationprotocol}, we re-implemented them under our training pipeline. For E2E-Spot and T-DEED, we leverage their publicly available code. For T-DEED's SGP-Mixer module, we adopt $B=2$ layers, a kernel size of $ks=9$, and a scalable factor of $r=4$, consistent with their SN-BAS experiments. For~\citet{santra2025precise}, no public code is available, so we implemented their proposed ASTRM module from scratch. All methods are trained on sequences of $L=100$ frames with a spatial resolution of $398\times 224$.

\subsection{Redundancy-aware methods}

\begin{figure}[t]
    \centering
    \includegraphics[width=\linewidth]{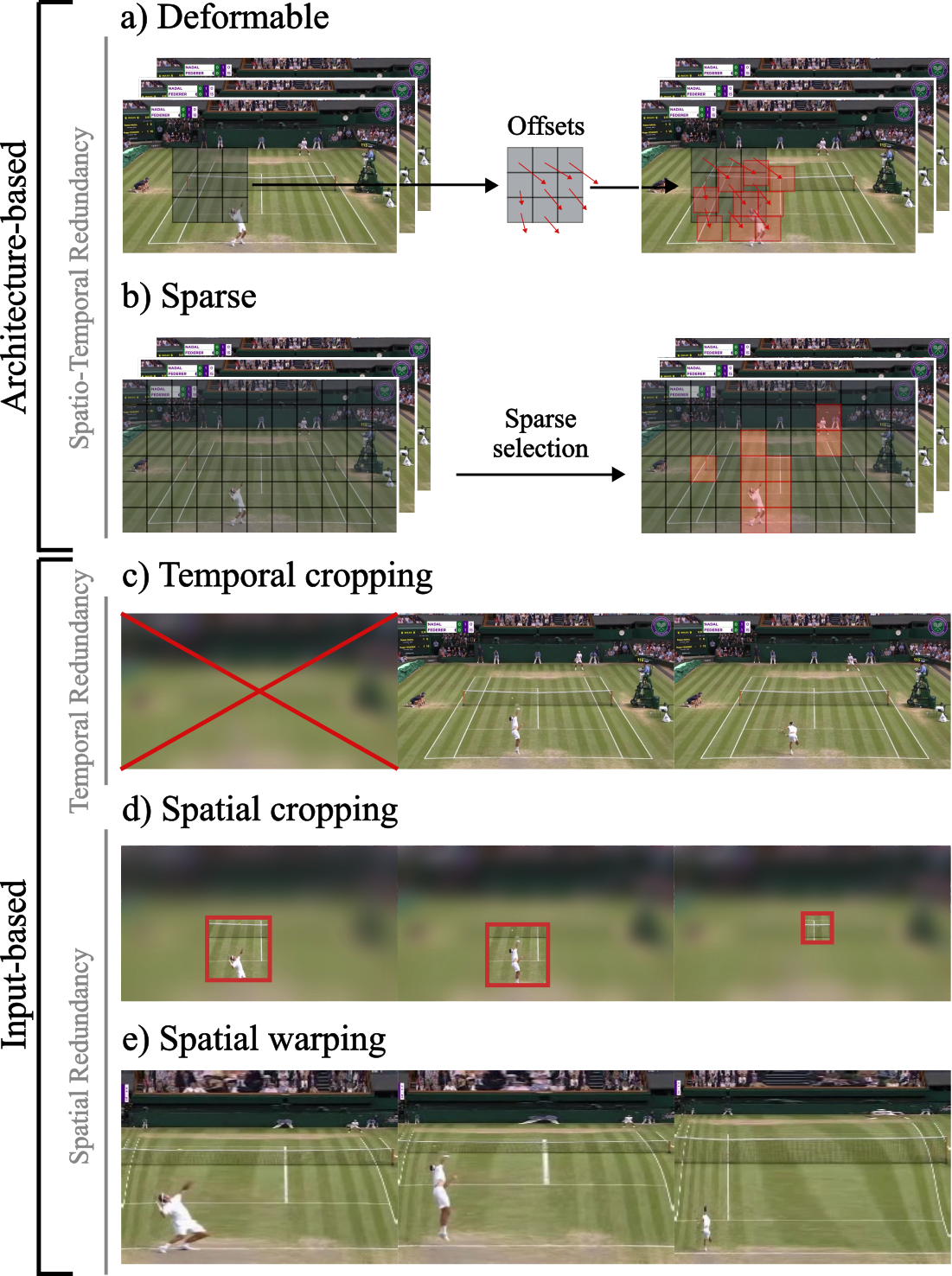}
    \caption{Illustration of the taxonomy of methods addressing spatio-temporal redundancy. We categorize approaches into \textit{architecture-based} and \textit{input-based}, and indicate whether each method handles spatial redundancy, temporal redundancy, or both.}
    \label{fig:rwfigure}
    \vspace{-0.5cm}
\end{figure}

We provide additional implementation details for the comparison of AdaSpot with alternative redundancy-aware approaches in Sec.~\textcolor{blue}{4.3} of the main paper. We first report results for \underline{AdaSpot} under three configurations with low-resolution inputs of $(W_l, H_l) = \tfrac{1}{4}(W_h, H_h)$, $(W_l, H_l) = \tfrac{3}{8}(W_h, H_h)$, and $(W_l, H_l) = \tfrac{1}{2}(W_h, H_h)$, as well as for a \underline{single low-resolution baseline} that uses only the low-resolution branch under the same input resolutions. For the redundancy-aware methods, we adopt the taxonomy shown in~\cref{fig:rwfigure}, which distinguishes architecture-based methods --those that mitigate redundancy at the feature level-- from input-based methods --those that address redundancy at the input level. Since AdaSpot targets spatial redundancy, which is more relevant for the PES task, we restrict our comparisons to methods that explicitly handle spatial redundancy. Specifically, we evaluate AdaSpot against: (i) \underline{deformable convolutions~\cite{dai2017deformable}}, applied spatially; (ii) sparse convolutions~\cite{liu2015sparse}, also applied spatially in two variants --one using saliency maps (\underline{Sparse-Saliency}) and one using learned gating mechanisms (\underline{Sparse-Learned}) to select sparsity locations; (iii) learnable pixel-space cropping (\underline{AdaFocus-v2~\cite{adafocus-v2}}); (iv) learnable feature-space cropping with variable size regions (\underline{Uni-AdaFocus~\cite{adafocus-v4}}); and (v) \underline{saliency-driven frame warping~\cite{liu2022task}}. Additional details for each approach are provided below. 

\mysection{Deformable convolutions.} For this approach, we adopt a simplified version of the AdaSpot architecture consisting of a single branch that processes frames at a fixed spatial resolution. Concretely, we retain one feature extractor, the temporal modeler, and the prediction head, while removing the RoI selector, the second feature extractor, the linear projectors, and the aggregation module. We then incorporate deformable convolutions into the remaining feature extractor. Specifically, all convolutions outside the initial ``stem'' block --so as to preserve standard dense early processing-- with kernel size larger than $1\times 1$ are replaced by deformable convolutions with matching configuration. We report results for two variations of this approach, corresponding to input spatial resolutions of $398\times224$ and $796\times 448$, which yield different computational costs.

\mysection{Sparse-Saliency.} This approach uses the same simplified architecture as the deformable-convolution baseline, but replaces the designated dense convolutions with sparse convolutions instead. The sparse activation pattern is determined using saliency maps: for each convolution, we compute a saliency map by channel-wise averaging the input features, following the procedure used in AdaSpot. We then retain the top $25\%$ of positions within each frame's feature maps with the highest activations as the active sparse locations. As in the deformable convolutions approach, we report results for two variants with input spatial resolutions of $398\times224$ and $796\times 448$.

\mysection{Sparse-Learned.} This approach is similar to Sparse-Saliency, but the sparse activation pattern is learned end-to-end using a lightweight gating module that predicts per-position importance scores via a linear layer followed by a sigmoid. We then select the top $25\%$ positions using a hard top-k during the forward pass, while employing a straight-through estimator (STE)~\cite{bengio2013estimating} in backpropagation to allow gradients to flow through the soft scores. The resulting masked features are then processed by the convolutional layer. As in the other methods, we evaluate two variants with input spatial resolutions of $398\times224$ and $796\times 448$.

\mysection{AdaFocus-v2.} For this approach, we adopt the same architecture as AdaSpot, replacing the RoI selector with the one proposed in~\citet{adafocus-v2}. Specifically, their RoI selector takes the feature maps $F_s$ as input and processes them through a series of spatial and temporal modules to produce per-frame predictions indicating the center of the region to crop. The approach is made differentiable via their learnable cropping mechanism, which incorporates a stop-gradient operation to improve training stability. We evaluate three variants of this method, corresponding to low-resolution inputs of $(W_l, H_l) = \tfrac{1}{4}(W_h, H_h)$, $(W_l, H_l) = \tfrac{3}{8}(W_h, H_h)$, and $(W_l, H_l) = \tfrac{1}{2}(W_h, H_h)$.

\mysection{Uni-AdaFocus.} This approach is analogous to the previous one, but replaces the RoI selector with the version proposed in~\citet{adafocus-v4}. Specifically, their method learns crop positions in the feature-space to improve training stability and is adapted to allow variable-size regions. As with the previous baseline, we evaluate three variants with low-resolution inputs of $(W_l, H_l) = \tfrac{1}{4}(W_h, H_h)$, $(W_l, H_l) = \tfrac{3}{8}(W_h, H_h)$, and $(W_l, H_l) = \tfrac{1}{2}(W_h, H_h)$.

\mysection{Saliency warping.} This approach uses the same AdaSpot architecture, but replaces the selected regions in the high-resolution branch with warped frames that emphasize the relevant regions, following~\citet{liu2022task}. We use the same saliency maps extracted for AdaSpot to guide the warping, as they provide reliable estimates of important regions. The warped frames are generated using the method proposed in~\citet{liu2022task}. As with other baselines, we evaluate three variants with low-resolution inputs of $(W_l, H_l) = \tfrac{1}{4}(W_h, H_h)$, $(W_l, H_l) = \tfrac{3}{8}(W_h, H_h)$, and $(W_l, H_l) = \tfrac{1}{2}(W_h, H_h)$.

\section{Additional ablation studies}
\label{sec:ablation}

In this section, we extend the ablation analysis presented in Sec.~\textcolor{blue}{4.3} of the main paper. Specifically, we first provide a more detailed examination of the components and parameters of our proposed AdaSpot approach. We then analyze the instability of AdaSpot compared to learnable-based alternatives, and finally, we further examine and discuss the selected RoIs for AdaSpot in comparison with those of alternative redundancy-aware methods that operate in the input space. 

\subsection{Extended component analysis}

We extend the component analysis from Sec.~\textcolor{blue}{4.3} by first providing a visual examination of the \textit{center bias} issue that arises when using zero-padding. We then report additional ablations on key components and parameters of AdaSpot, including alternative fusion strategies, different crop sizes, weight-sharing between the feature extractors of the low- and high-resolution branches, employing adaptive RoI aspect ratios, selecting multiple RoIs per frame, and analyzing the sensitivity of the $\tau$ parameter.

\begin{figure}[t]
    \centering
    \includegraphics[width=\linewidth]{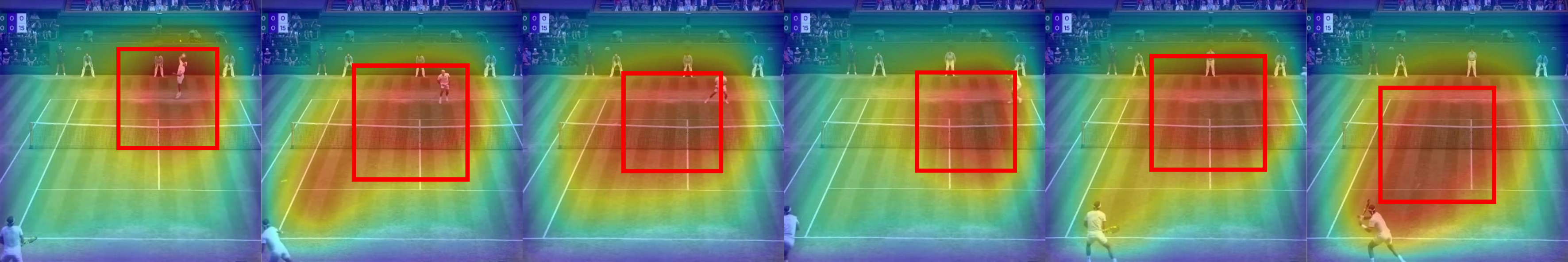}
    \caption{Qualitative visualization of saliency maps and the resulting selected RoIs when zero-padding is applied on the Tennis dataset. With zero-padding, the RoI selector ends up biased towards the central part of the frames.}
    \label{fig:zeropadding}
    \vspace{-0.5cm}
\end{figure}

\mysection{\textit{Center bias} extended analysis.} In Sec.~\textcolor{blue}{4.3} of the main paper, we reported a performance drop when replacing replicate padding with zero-padding. We attribute this drop to a \textit{center bias} introduced by zero-padding, which artificially reduces activation strength near the image borders~\cite{alsallakh2020mind}. ~\cref{fig:zeropadding} provides additional qualitative evidence for this effect by visualizing the resulting saliency maps and the RoIs selected when zero-padding is used. Although the saliency maps generally track the ball, activations near the borders are notably weaker than those at the center, leading the RoI selector to avoid choosing regions along the frame boundaries --even when the ball is located there. Consequently, the high-resolution branch receives less semantically meaningful crops, which negatively affects performance. This issue is most pronounced on the Tennis dataset and appears in certain runs, but when it arises, it can substantially degrade AdaSpot's effectiveness. In contrast, we do not observe any such behavior when using replicate padding, across all experiments and random seeds.

\begin{table}[t]
  \caption{Extended ablation study of AdaSpot components on Tennis and SN-BAS, evaluating the impact of alternative fusion mechanisms, crop sizes, backbone reuse for both low- and high-resolution branches, adaptive RoI aspect ratios, and multiple RoIs per frame.}
  \label{tab:additionalAblationsTask}
  \centering
  \resizebox{\columnwidth}{!}{
    \begin{tabular}{llc>{\columncolor[gray]{0.95}}c>{\columncolor[gray]{0.95}}cc>{\columncolor[gray]{0.95}}c}
      \toprule
      & & \multicolumn{3}{c}{Tennis} & \multicolumn{2}{c}{SN-BAS} \\
      \cmidrule(lr){3-5} \cmidrule(lr){6-7}
      \multicolumn{2}{l}{\textbf{Experiment}} & $\delta=0$f & $1$f & $2$f & $\delta=0.5$s & $1$s  \\
      \midrule
      \multicolumn{2}{l}{AdaSpot} & 73.30 & 96.90 & 97.47 & 53.02 & 56.43 \\
      \midrule
      (a) & \textit{Fusion mechanism} & & & & & \\
      & \quad mean & 71.26 & 96.48 & 97.07 & 50.55 & 54.56 \\
      & \quad product & 71.88 & 96.75 & 97.37 & 51.24 & 54.95 \\
      & \quad linear & 71.36 & 96.37 & 96.96 & 51.30 & 55.39 \\
      & \quad frame-gated & 71.86 & 96.27 & 96.80 & 52.03 & 56.25 \\
      & \quad channel-gated & 72.93 & 96.87 & 97.41 & 52.12 & 55.60 \\
      \midrule
      (b) & \textit{Crop size} & & & & &  \\
      & \quad $56\times 56$ & 71.05 & 96.52 & 97.18 & 51.53 & 55.61 \\
      & \quad $84\times 84$ & 72.19 & 96.66 & 97.28 & 51.16 & 55.26 \\
      & \quad $168\times 168$ & 72.45 & 96.89 & 97.47 & 52.08 & 55.58 \\
      & \quad $224\times 224$ & 73.02 & 96.77 & 97.28 & 50.60 & 54.66 \\
      \midrule
      (c) & \textit{Extractor reuse} & & & & &  \\
      & \quad yes & 71.70 & 96.52 & 97.19 & 51.90 & 56.04 \\
      \midrule
      (d) & \textit{RoI aspect ratio} & & & & &  \\
      & \quad adaptive & 71.66 & 96.62 & 97.24 & 51.61 & 54.97 \\
      \midrule
      (e) & \# RoIs per frame & & & & & \\
      & \quad 2 RoIs per frame & 72.06 & 96.83 & 97.35 & 49.19 & 52.70 \\
      \bottomrule
    \end{tabular}
  }
  \vspace{-0.5cm}
\end{table}

\mysection{Additional component and parameter ablations for AdaSpot.} In~\cref{tab:additionalAblationsTask}(a) we compare AdaSpot's max-based fusion of $F_l'$ and $F_h'$ with several \textbf{alternative fusion mechanisms}. Specifically, we evaluate: (i) \textit{mean} --the per-position element-wise average; (ii) \textit{product} --the element-wise (Hadamard) product; (iii) \textit{linear} --concatenating the feature vectors along the channel dimension and projecting back to dimension $d$ through a linear layer; (iv) \textit{frame-gated} --a per-frame gating mechanism that predicts a scalar $\alpha$ from the features and fuses them as $F_f = \alpha F_l' + (1-\alpha) F_h'$; and (v) \textit{channel-gated} --the same gating approach but predicting channel-wise gates instead of a single scalar. As shown by the results, none of these alternatives surpass the simple max-based aggregation, with several of them also incurring additional computational overhead. ~\cref{tab:additionalAblationsTask}(b) reports results for \textbf{varying crop sizes}. Reducing the crop below $112\times112$ slightly decreases performance, likely because smaller regions either capture less content or are downsampled to lower resolution when resized to $(W_r, H_r)$. Increasing the crop size yields results closer to the baseline but does not surpass it, which we attribute to larger RoIs introducing extra context that is not task-relevant. ~\cref{tab:additionalAblationsTask}(c) shows that \textbf{reusing extractor parameters} for both the low- and high-resolution branches still achieves strong performance with only minor drops. This shows that AdaSpot can be made more parameter-efficient, reducing total parameters by $37\%$ while decreasing the strictest metrics by only $-1.60$ and $-1.12$ on Tennis and SN-BAS, respectively. Additionally, ~\cref{tab:additionalAblationsTask}(d) evaluates \textbf{adaptive RoI aspect ratios}, where the RoI is no longer constrained to a fixed aspect ratio. Instead, here we take the rectangular region according to the saliency spread without enforcing this constraint. This modification results in a performance drop of $-1.64$ and $-1.41$ on Tennis and SN-BAS, respectively. We attribute this to the increased complexity of modeling RoIs with varying aspect ratios, which complicates training and reduces performance. In~\cref{tab:additionalAblationsTask}(e) we evaluate using \textbf{multiple RoIs per frame}. We extend AdaSpot to the multi-RoI setting by selecting a second region corresponding to the highest remaining saliency after excluding the first RoI for each frame. This results in two RoI clips that are processed through the shared high-resolution extractor and aggregated using element-wise maximum. For simplicity, a fixed region size is used in these experiments. The results show that incorporating more than one RoI consistently degrades performance, indicating that additional regions do not provide complementary information and instead introduce noise. This finding aligns with our qualitative analysis (Sec.~\textcolor{blue}{4.4}), where saliency maps typically highlight a single dominant region, suggesting that a single RoI suffices for current PES benchmarks. While multi-RoI modeling could benefit scenarios with multiple simultaneous events, such dynamics are not present in the standard PES datasets. A more extensive study of multi-RoI extensions of AdaSpot, evaluated on datasets with concurrent events and multiple relevant regions, is therefore left for future work. 
\begin{figure}[t]
    \centering
    \includegraphics[width=\linewidth]{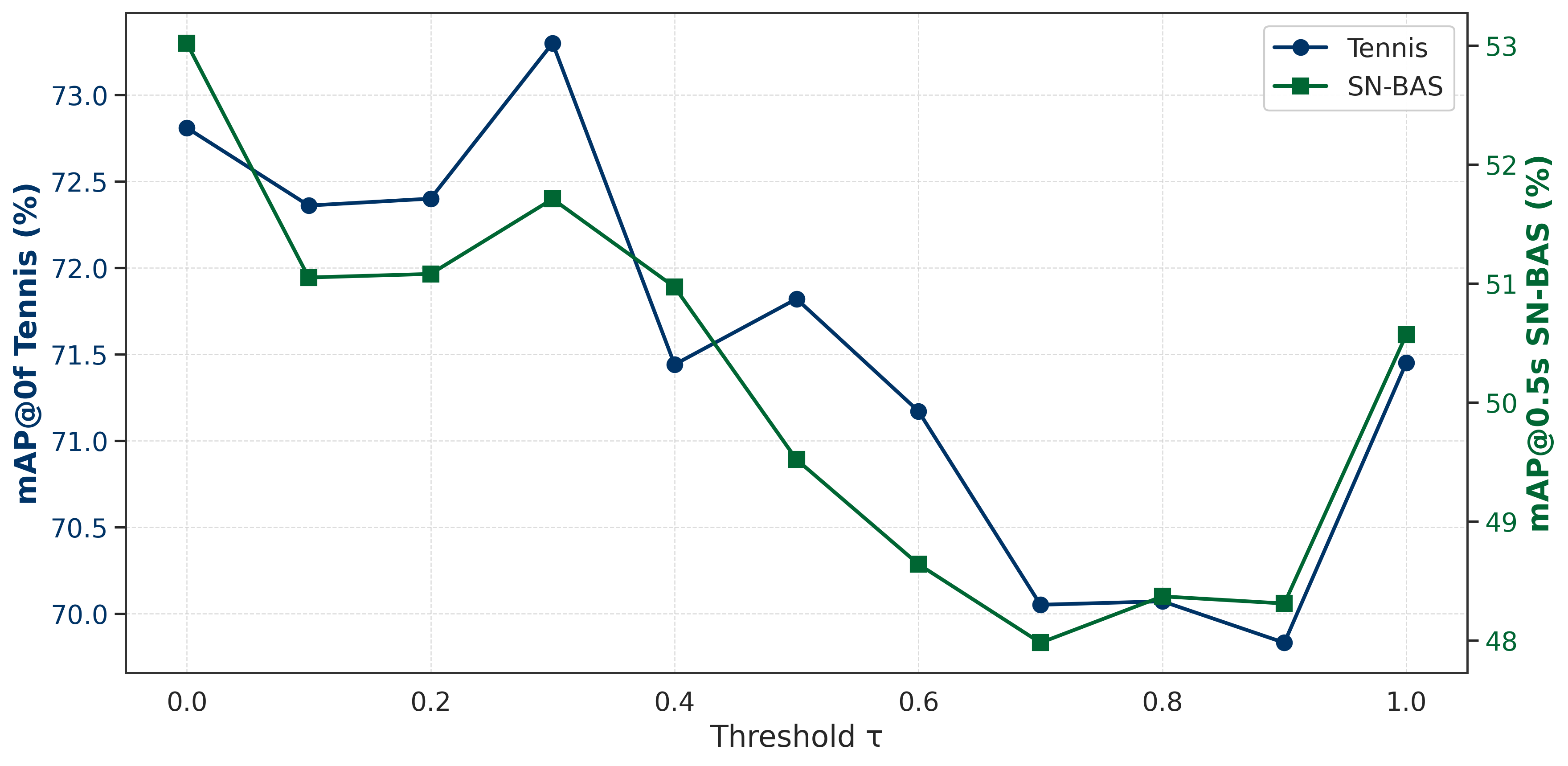}
    \caption{Sensitivity of AdaSpot performance ($\text{mAP}@0\text{f}$ / $\text{mAP}@0.5\text{s}$) to variations in the threshold parameter $\tau$. The blue line denotes the Tennis dataset (left y-axis), while the green line denotes the SN-BAS dataset (right y-axis).}
    \label{fig:threshold}
    \vspace{-0.2cm}
\end{figure}
Finally, ~\cref{fig:threshold} analyzes the sensitivity to the \textbf{threshold parameter $\tau$}. Both datasets exhibit similar trends, with two main performance peaks, at $\tau=0$ and around $\tau=0.3$. The peak at $\tau=0$ arises from using fixed-size RoIs, which simplifies modeling in the high-resolution branch despite occasionally omitting context. As $\tau$ increases, performance initially decreases, indicating that the added contextual information does not compensate for the difficulty of modeling variable-size RoIs. Around $\tau = 0.3$, the added context becomes beneficial enough to counteract this effect, producing the second peak. Beyond this range, performance declines once RoIs include excessive non-informative content while still varying in size. At $\tau=1$, we observe a small final peak, as the RoI becomes the full frame, again yielding fixed-size regions that simplify modeling --though at the cost of negating the purpose of the high-resolution branch, which now processes downsampled full-view clips.

\subsection{Instability analysis of learnable cropping}

\cref{tab:instability} presents a comparative instability analysis of AdaSpot against alternative learnable cropping methods: AdaFocus-v2 (AF-v2) and Uni-AdaFocus (Uni-AF). Across datasets, AdaSpot consistently achieves lower standard deviation, indicating more stable training. AF-v2 exhibits high variability, which is partially mitigated in Uni-AF. In addition, AdaSpot demonstrates more robust RoI selection, as reflected by higher performance when using high-resolution features only. In contrast, AF-v2 and Uni-AF produce more failure cases. These results highlight AdaSpot's improved training stability and RoI robustness.

\begin{table}[t]
  \centering
  \caption{Comparison of instability under the strictest metric for different learnable cropping methods with AdaSpot. Bold indicates best; (mean\std{std.} across 3 runs).}
    \centering
    \label{tab:instability}
    \resizebox{\columnwidth}{!}{
      \begin{tabular}{lccc|ccc}
      \toprule
        & \multicolumn{3}{c}{Low \& high-res features}
        & \multicolumn{3}{c}{High-res features only} \\
        \midrule
        \textbf{Dataset} & AF-v2 & Uni-AF & AdaSpot & AF-v2 & Uni-AF & AdaSpot \\
        \midrule
        Tennis & 70.6\std{1.5} & 70.2\std{1.0} & \textbf{73.3}\std{0.5} & 68.8\std{0.9} & 65.9\std{0.9} & \textbf{71.9}\std{0.4} \\
        SN-BAS & 49.0\std{2.0} & 49.6\std{1.3} & \textbf{53.0}\std{0.5} & 23.1\std{15.1} & 39.4\std{3.3} & \textbf{52.1}\std{0.7} \\
        \bottomrule
      \end{tabular}
    }
  \vspace{-0.5cm}
\end{table}

\subsection{Qualitative RoI comparison}

\begin{figure}[t]
    \centering
    \includegraphics[width=\linewidth]{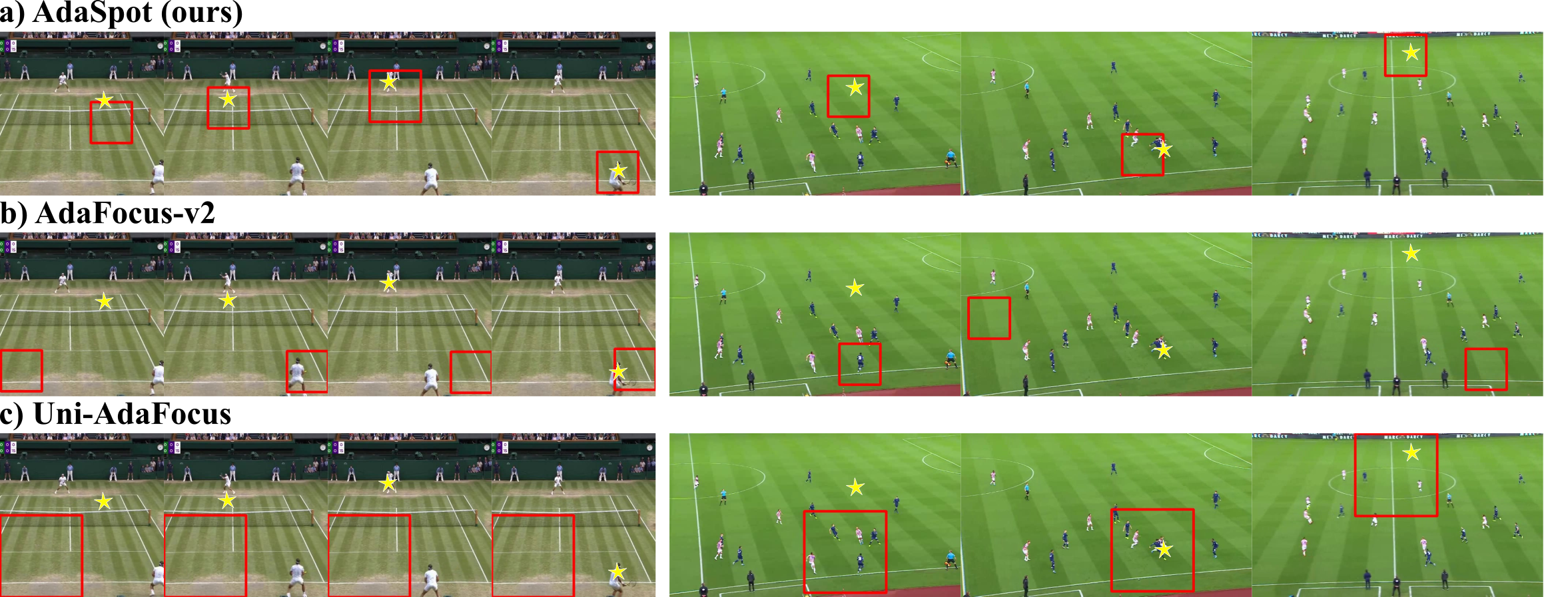}
    \caption{Qualitative comparison of the RoIs selected by AdaSpot, AdaFocus-v2, and Uni-AdaFocus on the Tennis (left) and SN-BAS (right) datasets. For visualization, we mark the ball position in each frame with a star, as actions in these datsets occur around the ball; thus, relevant RoIs should contain or closely surround it.}
    \label{fig:qC}
    \vspace{-0.5cm}
\end{figure}

\cref{fig:qC} compares the RoIs selected by our AdaSpot with those produced by input-based alternatives, specifically AdaFocus-v2 and Uni-AdaFocus. As shown, these alternative methods frequently fail to capture task-relevant regions, introducing noise during training and diminishing the effectiveness of the high-resolution branch --ultimately leading to the performance drops reported in Sec.~\textcolor{blue}{4.3} of the main paper. On Tennis, both AdaFocus-v2 and Uni-AdaFocus tend to converge to largely static corner crops, likely because some actions commonly occur near those areas. On SN-BAS, the crops move more dynamically, and Uni-AdaFocus localizes relevant regions more reliably (\eg, around the ball). However, its adaptive region size often saturates to the maximum allowed area, causing fine-grained details to be lost after resizing to $(W_r, H_r)$. In contrast, AdaSpot consistently selects stable, semantically meaningful RoIs. As discussed in the main paper, we attribute the limitations of such learnable-cropping approaches to the training instabilities identified in prior work~\cite{adafocus-v4}, which our training-free RoI selector inherently avoids. 

\section{Efficiency analysis}
\label{sec:efficiency}

In this section, we extend the efficiency analysis presented in Sec.~\textcolor{blue}{4.2} of the main paper. \cref{tab:cost} reports the number of parameters and GFLOPs required to process a single clip under both the PES setting (base resolution $224{\times}224$) and the ES setting (base resolution $398{\times}224$). For ~\citet{santra2025precise}, these values are derived from our re-implementation of their ASTRM module, which may therefore differ slightly from those originally reported. We exclude UGLF~\cite{tran2024unifying} from the comparison due to missing details regarding their vision-language module in the released code. While E2E-Spot$_{200\text{MF}}$, T-DEED$_{200\text{MF}}$, and ~\citet{santra2025precise} all employ RegNetY-200MF as their base extractor, E2E-Spot stands out among the most efficient in both parameter count and GFLOPs. T-DEED exhibits comparable computational cost but is substantially more parameter-intensive due to its SGP-Mixer module used for temporal modeling. In contrast, ~\citet{santra2025precise} maintains a low parameter count but incurs higher GFLOPs because the ASTRM module is inserted early in the backbone, where feature maps are still high resolution, thereby increasing the overall computational cost. AdaSpot$^s$, which uses the same base extractor, introduces only a marginal increase in parameters and GFLOPs relative to E2E-Spot$_{200\text{MF}}$. The additional parameters arise from duplicating the extractor for the high-resolution branch, while the extra computation stems from processing the RoI clips through this branch --adding approximately 6 GFLOPs for our standard $112\times 112$ RoI configuration. This small overhead enables AdaSpot to preserve fine-grained details and yields substantial performance improvements (see Sec.~\textcolor{blue}{4.2} of the main paper), resulting in a stronger efficiency-accuracy trade-off. When comparing larger extractor configurations --E2E-Spot$_{800\text{MF}}$, T-DEED$_{800\text{MF}}$, and AdaSpot$^b$-- we observe that AdaSpot$^b$, despite using a smaller backbone (RegNetY-400MF) and thus being more efficient, still achieves state-of-the-art performance across both PES and ES datasets (see Sec.~\textcolor{blue}{4.2} of the main paper). Additionally, for AdaSpot, inference on a single clip requires only 1.97GB of GPU memory, enabling inference even on small GPUs.

\begin{table}[t]
  \caption{Efficiency comparison of AdaSpot with state-of-the-art methods in both the PES setting (typically using $224{\times}224$ inputs) and the ES setting (using $398{\times}224$ inputs). For each configuration, we report the number of parameters (in millions) and the computational cost in GFLOPs.}
  \label{tab:cost}
  \centering
  \resizebox{\linewidth}{!}{
  \begin{tabular}{lcccc}
    \toprule
   & \multicolumn{2}{c}{PES} & \multicolumn{2}{c}{ES} \\
   \cmidrule(lr){2-3} \cmidrule(lr){4-5} 
   
     Model & P(M) & GFLOPs & P(M) & GFLOPs \\
    \midrule

    E2E-Spot$_{200\text{MF}}$~\cite{hong2022spotting} & 4.49 & 23.13 & 4.49 & 40.78  \\
    E2E-Spot$_{800\text{MF}}$~\cite{hong2022spotting} & 12.70 & 84.93 & 12.70 & 150.02 \\

    T-DEED$_{200\text{MF}}$~\cite{xarlest} & 16.42 & 21.97 & 12.31 & 39.58 \\
    T-DEED$_{800\text{MF}}$~\cite{xarlest} & 64.26 & 86.34 & 46.22 & 151.31 \\

    Santra et al.~\cite{santra2025precise} & 6.46 & 57.84 & 6.84 & 82.51 \\
    
    \midrule
    
    \textbf{AdaSpot$^s$} & 7.58 & 29.78 & 7.58 & 46.18 \\
    \textbf{AdaSpot$^b$} & 10.63 & 36.78 & 10.63 & 90.04 \\
    \bottomrule
    \end{tabular}
    }
  \vspace{-0.5cm}
\end{table}

\section{Randomness analysis}
\label{sec:randomness}

Training deep neural networks involves multiple sources of randomness (\eg, data sampling, weight initialization, and data augmentation), which can lead to noticeable performance variability across runs. Despite this, most PES methods report results from a single training run, due to the substantial computational cost of these pipelines. This practice can produce benchmarks that are sensitive to run-to-run fluctuations, making claims difficult to verify or reproduce. To provide more robust evaluations, we report results over three runs using different random seeds. For all experiments, we report the mean performance, and for the main results, we additionally provide the standard deviation to reflect variability across runs. While more runs would allow more rigorous statistical analysis --three runs are insufficient for reliable significance testing-- the high computational demands of PES frameworks make extensive multi-run evaluation impractical. Nevertheless, our three-run reporting offers improved robustness over the single-run convention used in prior work. 

\section{Post-processing analysis}
\label{sec:postprocessing}

In this section, we analyze the sensitivity of PES methods to the choice of post-processing. \cref{tab:postprocessing} presents AdaSpot results on the Tennis dataset under different post-processing configurations. Specifically, we compare standard Non-Maximum Suppression (NMS)~\cite{neubeck2006efficient}, and Soft Non-Maximum Suppression (Soft-NMS)~\cite{bodla2017soft}, evaluating multiple window sizes $\omega \in \{1, 2, 3, 4, 5\}$. As shown, Soft-NMS generally outperforms NMS for the strictest metric (mAP$@0$f) while achieving comparable results for looser metrics. Within Soft-NMS, smaller window sizes slightly favor stricter metrics, whereas larger windows benefit more relaxed metrics. Following prior work~\cite{xarles2024t, santra2025precise}, we adopt a configuration that balances performance across all tolerances and use Soft-NMS with $\omega=2$. For ES experiments, where more relaxed evaluation protocols are used, larger window sizes are preferable; in this case, we find $\omega=12$ to offer the best trade-off. To ensure fair comparisons with state-of-the-art methods (Sec.~\textcolor{blue}{4.2} main paper), whenever possible, all results are re-extracted using the same post-processing settings.

\begin{table}[t]
  \caption{Post-processing sensitivity analysis on the Tennis dataset. We report results for standard NMS and Soft-NMS using different window sizes $\omega$. Bold and underlined values indicate the best and second-best results.}
  \label{tab:postprocessing}
  \centering
  \resizebox{\columnwidth}{!}{
    \begin{tabular}{llc>{\columncolor[gray]{0.95}}c>{\columncolor[gray]{0.95}}c}
      \toprule
      & & \multicolumn{3}{c}{Tennis} \\
      \cmidrule(lr){3-5}
      \multicolumn{2}{l}{\textbf{Post-processing}} & $\delta=0$f & $1$f & $2$f  \\
      \midrule
      \multirow{5}{*}{NMS~\cite{neubeck2006efficient}} 
        & $\omega=1$ & 62.82 & \textbf{96.93} & \underline{97.61} \\
        & $\omega=2$ & 62.45 & 96.61 & \textbf{97.64} \\
        & $\omega=3$ & 62.35 & 96.11 & 97.58 \\
        & $\omega=4$ & 62.32 & 95.94 & 97.47 \\
        & $\omega=5$ & 62.29 & 95.84 & 97.29 \\
      \midrule
      \multirow{5}{*}{Soft-NMS~\cite{bodla2017soft}} 
        & $\omega=1$ & \textbf{75.05} & 96.02 & 96.50 \\
        & $\omega=2$ & \underline{73.30} & 96.90 & 97.47 \\
        & $\omega=3$ & 71.53 & \underline{96.92} & 97.56 \\
        & $\omega=4$ & 70.35 & 96.81 & 97.60 \\
        & $\omega=5$ & 69.36 & 96.70 & 97.59 \\
      \bottomrule
    \end{tabular}
  }
  \vspace{-0.5cm}
\end{table}

\section{Additional results and visualizations}
\label{sec:additional_results}

In this section, we first analyze the per-class performance of AdaSpot in comparison with other state-of-the-art methods (E2E-Spot and T-DEED), and provide an approximate per-class evaluation of the RoI selection. We then present additional results, including F3Set evaluations under their proposed metrics, as well as visualizations of the generated saliency maps, the selected RoIs, and the corresponding model predictions for AdaSpot.

\subsection{Per-class results}

Here, we report per-class results of AdaSpot compared with E2E-Spot and T-DEED, using the best-performing version of each model for each dataset. On the \textbf{Tennis} (\cref{tab:tennis}) and \textbf{FineDiving} (\cref{tab:finediving}) datasets, we observe the same trend as in the aggregated results from the main paper, with AdaSpot outperforming the other two methods across all event classes. In Tennis, the most notable improvements over E2E-Spot occur on ``far-court swings'' and ``far-court serves'', highlighting that AdaSpot is particularly effective for far-view actions where uniform resolution downsampling can hinder performance. By focusing higher-resolution attention on relevant regions, AdaSpot better captures these challenging events. On \textbf{FineGym} (\cref{tab:finegym}), the performance across methods is generally similar, but AdaSpot maintains competitive results across all classes, achieving strong overall performance. Finally, on \textbf{SN-BAS} (\cref{tab:snbas}), AdaSpot again demonstrates superiority, achieving the best results for all but two classes. These results confirm that the improvements introduced by AdaSpot are consistent across most event categories, reinforcing its general effectiveness.

\begin{table}[t]
  \caption{Per-class analysis on the Tennis dataset. For each event class, we report the total number of observations and the AP$@0$f results for the best-performing versions of E2E-Spot, T-DEED, and AdaSpot, as well as for an AdaSpot variant using high-resolution features only (HR-only). Event classes are sorted in descending order of observations. The best result per class is highlighted in bold, and the second-best is underlined.}
  \label{tab:tennis}
  \centering
  \resizebox{\columnwidth}{!}{
     \begin{tabular}{lcccc|c} \toprule
     &  & \multicolumn{4}{c}{AP ($\delta=0$f)} \\
     \cmidrule(lr){3-6}
      Event & Nº observations & E2E-Spot & T-DEED & \textbf{AdaSpot} & HR-only \\
      \midrule
      Far-court ball bounce & 8150 & \underline{76.91} & 59.47 & \textbf{77.20} & 75.38 \\
      Near-court ball bounce  & 8127 & \underline{76.79} & 68.19 & \textbf{78.98} & 78.91 \\
      Far-court swing  & 7123 & \underline{53.24} & 41.52 & \textbf{64.76} & 58.77 \\
      Near-court swing  & 7044 & \underline{56.42} & 48.85 & \textbf{58.83} & 58.49 \\
      Near-court serve  & 1690 & \underline{76.79} & 67.91 & \textbf{79.24} & 78.67 \\
      Far-court serve & 1657 & \underline{80.10} & 64.66 & \textbf{85.09} & 81.95 \\
      
      \bottomrule
    \end{tabular}
  }
\end{table}

\begin{table}[t]
  \caption{Per-class analysis on the FineDiving dataset. For each event class, we report the total number of observations and the AP$@0$f results for the best-performing versions of E2E-Spot, T-DEED, and AdaSpot. Event classes are sorted in descending order of observations. The best result per class is highlighted in bold, and the second-best is underlined.}
  \label{tab:finediving}
  \centering
  \resizebox{\columnwidth}{!}{
     \begin{tabular}{lcccc} \toprule
     &  & \multicolumn{3}{c}{AP ($\delta=0$f)} \\
     \cmidrule(lr){3-5}
      Event & Nº observations & E2E-Spot & T-DEED & \textbf{AdaSpot} \\
      \midrule
      Entry & 2984 & 22.51 & \underline{24.02} & \textbf{26.74} \\
      Som(s).Pike  & 2152 & \underline{27.21} & 23.14 & \textbf{27.58} \\
      Som(s).Tuck  & 1071 & \underline{31.70} & 21.72 & \textbf{32.23} \\
      Twist(s)  & 803 & \underline{18.60} & 16.45 & \textbf{22.51} \\
      
      \bottomrule
    \end{tabular}
  }
  \vspace{-0.5cm}
\end{table}

\begin{table}[t]
  \caption{Per-class analysis on the FineGym dataset. For each event class, we report the total number of observations and the AP$@0$f results for the best-performing versions of E2E-Spot, T-DEED, and AdaSpot. Event classes are sorted in descending order of observations. The best result per class is highlighted in bold, and the second-best is underlined.}
  \label{tab:finegym}
  \centering
  \resizebox{\columnwidth}{!}{
     \begin{tabular}{lcccc} \toprule
     &  & \multicolumn{3}{c}{AP ($\delta=0$f)} \\
     \cmidrule(lr){3-5}
      Event & Nº observations & E2E-Spot & T-DEED & \textbf{AdaSpot} \\
      \midrule
      Uneven bars circles start & 6612 & \textbf{11.32} & 10.26 & \underline{10.28} \\
      Uneven bars circles end & 6612 & \textbf{20.19} & \underline{19.89} & 19.63\\
    Balance beam leap\_jump\_hop start & 4787 & 17.52 & \textbf{19.72} & \underline{18.07} \\
    Balance beam leap\_jump\_hop end & 4787 & 10.31 & \textbf{12.64} & \underline{10.81} \\
    Balance beam flight\_salto start & 4187 & 19.84 & \underline{22.72} & \textbf{24.35} \\
    Balance beam flight\_salto end & 4187 & 6.86 & \textbf{7.76} & \underline{7.48} \\
    Uneven bars transition\_flight start & 3389 & \textbf{29.86} & \underline{29.60} & 26.65 \\
    Uneven bars transition\_flight end & 3389 & \textbf{30.73} & 26.99 & \underline{28.09} \\
    Floor exercise leap\_jump\_hop start & 3238 & \textbf{27.32} & \underline{26.14} & 25.43 \\
    Floor exercise leap\_jump\_hop end & 3238 & \textbf{16.41} & 14.10 & \underline{14.91} \\
    Floor exercise back\_salto start & 2978 & \textbf{35.88} & 33.26 & \underline{34.86} \\
    Floor exercise back\_salto end & 2978 & \textbf{13.61} & 11.95 & \underline{12.83}\\
    Balance beam flight\_handspring start & 2893 & 17.64 & \textbf{19.93} & \underline{19.08} \\
    Balance beam flight\_handspring end & 2893 & 23.91 & \textbf{28.80} & \underline{26.50} \\
    Vault (timestamp 0) & 2031 & \textbf{2.53} & 1.90 & \underline{2.36} \\
    Vault (timestamp 1) & 2031 & \textbf{22.54} & \underline{22.53} & 20.15\\
    Vault (timestamp 2) & 2031 & 35.28 & \underline{39.90} & \textbf{41.80} \\
    Vault (timestamp 3) & 2031 & 5.43 & \textbf{7.07} & \underline{6.29} \\
    Uneven bars flight\_same\_bar start & 1624 & \underline{27.30} & \textbf{27.85} & 25.63 \\
    Uneven bars flight\_same\_bar end & 1624 & 26.50 & \textbf{27.58} & \underline{26.82} \\
    Balance beam turns start & 1371 & \underline{12.47} & \textbf{13.98} & 11.56 \\
    Balance beam turns end & 1371 & \underline{4.67} & \textbf{5.33} & 4.64 \\
    Floor exercise from\_salto start & 1345 & 26.48 & \underline{26.83} & \textbf{29.60} \\
    Floor exercise from\_salto end & 1345 & \textbf{8.97} & 8.52 & \underline{8.70} \\
    Uneven bars dismounts start & 1227 & \textbf{34.37} & \underline{33.50} & 33.03 \\
    Uneven bars dismounts end & 1227 & \textbf{10.65} & \underline{8.55} & 7.80 \\
    Balance beam dismounts start & 1218 & 21.70 & \textbf{34.94} & \underline{27.86} \\
    Balance beam dismounts end & 1218 & \textbf{7.11} & \underline{6.89} & 4.96 \\
    Floor exercise turns start & 1103 & 9.07 & \textbf{12.53} & \underline{11.41} \\
    Floor exercise turns end & 1103 & 11.07 & \textbf{15.30} & \underline{13.52} \\
    Floor exercise side\_salto start & 49 & \textbf{22.35} & 8.49 & \underline{19.80} \\
    Floor exercise side\_salto end & 49 & \underline{2.86} & 1.59 & \textbf{7.70} \\
      
      \bottomrule
    \end{tabular}
  }
\end{table}

\begin{table}[t]
  \caption{Per-class analysis on the SN-BAS dataset. For each event class, we report the total number of observations and the AP$@0.5$s results for the best-performing versions of E2E-Spot, T-DEED, and AdaSpot, as well as for an AdaSpot variant using high-resolution features only (HR-only). Event classes are sorted in descending order of observations. The best result per class is highlighted in bold, and the second-best is underlined.}
  \label{tab:snbas}
  \centering
  \resizebox{\columnwidth}{!}{
     \begin{tabular}{lcccc|c} \toprule
     &  & \multicolumn{4}{c}{AP ($\delta=0.5$s)} \\
     \cmidrule(lr){3-6}
      Event & Nº observations & E2E-Spot & T-DEED & \textbf{AdaSpot} & HR-only \\
      \midrule
       Pass & 4985 & \underline{85.15} & 83.44 & \textbf{85.94} & 85.11 \\
       Drive & 4300 & \underline{81.55} & 77.25 & \textbf{81.77} & 81.50 \\
        High pass & 761 & \textbf{79.30} & 76.33 & \underline{78.68} & 75.18 \\
        Header & 713 & \underline{68.14} & 54.27 & \textbf{68.87} & 62.62 \\
        Ball out of play & 551 & 16.97 & \underline{19.79} & \textbf{23.01} & 21.15 \\
        Throw-in & 362 & \underline{67.74} & 58.48 & \textbf{70.52} & 65.40 \\
        Cross & 261 & \underline{64.27} & 62.64 & \textbf{69.66} & 52.65 \\
        Ball player block & 223 & \underline{24.94} & 16.46 & \textbf{24.98} & 21.34 \\
        Shot & 169 & \underline{51.23} & 44.31 & \textbf{55.21} & 53.21 \\
        Player successful tackle & 74 & \textbf{5.64} & 0.92 & \underline{3.77} & 1.84 \\
      
      \bottomrule
    \end{tabular}
  }
  \vspace{-0.5cm}
\end{table}

\subsection{Per-class RoI analysis}

Per-class RoI analysis is limited by the lack of ground-truth RoIs. However, evaluating an AdaSpot variant that uses only high-resolution features provides an approximate measure of RoI precision for each event class. We report these values for Tennis and SN-BAS in ~\cref{tab:tennis} and ~\cref{tab:snbas}. As shown in the tables, in Tennis, the largest performance drops compared to the full AdaSpot model occur for far-court events, indicating less precise RoI selection for distant actions. In contrast, close-court events show performance near the baseline, suggesting accurate RoI selection for nearby events. For SN-BAS, the most pronounced effect is observed for the cross event, which depends not only on the player interacting with the ball but also on the broader context of where the ball is headed, which is not covered by the RoI. 

\subsection{F3Set additional evaluation}

\cref{tab:resultsF3Set} presents a further evaluation on the F3Set dataset. Both AdaSpot variants outperform F$^3$ED across all mAP metrics and the F1 score, with substantial gains. However, on the Edit score, AdaSpot performs slightly lower, highlighting the contribution of the additional context refinement module introduced in F$^3$ED. Overall, these results demonstrate that AdaSpot achieves strong performance even on the more fine-grained event classes featured in F3Set.

\begin{table}[t]
  \centering
  \caption{Comparison of AdaSpot with F$^3$ED on the F3Set dataset using standard PES mAP metrics, as well as F1 and Edit scores. Results show the mean over three random seeds with the corresponding standard deviation (\std). Bold and underlined values indicate the best and second-best results.}
  \label{tab:f3set}

    \centering
    \label{tab:resultsF3Set}
    \resizebox{\columnwidth}{!}{
      \begin{tabular}{lccccc}
      \toprule
        & mAP$@0$f & mAP$@1$f & mAP$@2$f & F1$_{evt}$ & Edit \\
        \midrule
        F$^3$ED~\cite{liu2025f} & 24.8 & 60.7 & 64.8 & 40.3 & \textbf{74.0} \\
        \midrule
        \textbf{AdaSpot$^s$} & \underline{53.55\std{1.2}} & \underline{67.76\std{0.8}} & \underline{68.41\std{1.0}} & \underline{48.8\std{1.1}} & 72.6\std{0.4} \\
        \textbf{AdaSpot$^b$} & \textbf{55.38\std{0.3}} & \textbf{69.37\std{0.2}} & \textbf{69.94\std{0.2}} & \textbf{51.66\std{0.6}} & \underline{73.66}\std{0.4} \\
        \bottomrule
      \end{tabular}
    }
  \vspace{-0.5cm}
\end{table}

\subsection{Qualitative results}

\mysection{Saliency maps and selected RoIs.} To complement the visualizations in Sec.~\textcolor{blue}{4.4}, we provide in the Supplementary Material two example clips per dataset corresponding to the best-performing AdaSpot version, showing both the saliency maps and the selected RoIs. In \textbf{Tennis} (\textit{Video\_SaliencyRoIs\_Tennis\_1.mp4} and \textit{Video\_SaliencyRoIs\_Tennis\_2.mp4}), as previously discussed, events revolve around the ball. We observe that, in most frames, the areas of highest saliency --and consequently the selected RoIs-- align closely with the ball’s position. In a few cases, such as when the ball is in the air with multiple frames before or after an action, saliency occasionally shifts toward the players. However, as the clip progresses and approaches an action, the saliency consistently returns to the ball. Additionally, the generated RoIs move smoothly across frames, which facilitates effective spatio-temporal modeling within the high-resolution extractor. In \textbf{FineDiving} (\textit{Video\_SaliencyRoIs\_FineDiving\_1.mp4} and \textit{Video\_SaliencyRoIs\_FineDiving\_2.mp4}), where events center on a single athlete performing a dive, the RoIs consistently capture the athlete in nearly all frames while moving smoothly throughout the clip, demonstrating robust and reliable RoI localization. In \textbf{FineGym} (\textit{Video\_SaliencyRoIs\_FineGym\_1.mp4} and \textit{Video\_SaliencyRoIs\_FineGym\_2.mp4}), events again focus on a single athlete, and the saliency maps reliably highlight regions including the athlete. However, the camera views in this dataset are more varied, with some closer shots resulting in RoIs that cover only part of the athlete. In these cases, the selected regions tend to focus on the most relevant parts for the event (\eg, the hands contacting the vault during a vault, or the feet and floor when landing from a jump). We hypothesize that such closer views may explain why AdaSpot achieves slightly more modest results on this dataset, as the downsampled full-view frames already contain much of the necessary fine-grained detail. In \textbf{F3Set} (\textit{Video\_SaliencyRoIs\_F3Set\_1.mp4} and \textit{Video\_SaliencyRoIs\_F3Set\_2.mp4}), which resembles the Tennis dataset, we observe similar patterns: the highest saliency and selected RoIs closely align with the ball's position. Finally, in \textbf{SN-BAS} (\textit{Video\_SaliencyRoIs\_SNBAS\_1.mp4} and \textit{Video\_SaliencyRoIs\_SNBAS\_2.mp4}), events are again ball-centric. In both clips, the saliency maps and selected RoIs consistently follow the ball, producing semantically meaningful regions. Only in frames without nearby events, saliency spreads more evenly across the scene, occasionally resulting in the ball falling outside the RoI.

\mysection{AdaSpot predictions.} We additionally provide, in the Supplementary Material, one example clip per dataset showing AdaSpot's predictions, together with the temporal distance errors relative to the corresponding ground-truth annotations. In \textbf{Tennis} (\textit{Video\_Predictins\_Tennis.mp4}), the strong performance reported in Tab.~\textcolor{blue}{1} of the main paper is clearly reflected visually: all actions are detected with high temporal precision. In \textbf{FineDiving} (\textit{Video\_Predictins\_FineDiving.mp4}), all actions are still correctly identified, although some exhibit larger temporal localization errors. In \textbf{FineGym} (\textit{Video\_Predictins\_FineGym.mp4}), we observe occasionally multiple predictions around a single ground-truth event, which stem from the ambiguity in localizing certain event types. Finally, in \textbf{SN-BAS} (\textit{Video\_Predictins\_SNBAS.mp4}), predictions generally follow the ground truth well, achieving good precision under the more relaxed ES evaluation setting, with only one missed action near the end of the clip.

\end{document}